\documentclass[11pt]{article}

\usepackage[utf8]{inputenc}
\usepackage[T1]{fontenc}
\usepackage{microtype}
\usepackage[margin=1in]{geometry}
\usepackage{amsmath,amssymb}
\usepackage{booktabs}
\usepackage{array}
\usepackage{graphicx}
\usepackage{tikz}
\usetikzlibrary{positioning,arrows.meta,fit,backgrounds,calc}
\usepackage[ruled,vlined]{algorithm2e}
\usepackage{placeins}  
\usepackage{xcolor}
\usepackage{listings}
\usepackage{enumitem}
\usepackage{caption}
\usepackage[hidelinks]{hyperref}

\graphicspath{{figs/}{figures/}{../figures/}}

\definecolor{kw}{rgb}{0.10,0.25,0.55}
\definecolor{cm}{rgb}{0.30,0.45,0.30}
\newcommand{\code}[1]{\texttt{\small #1}}
\title{\textbf{Execution-State Capsules}\\[2pt]
\large Graph-Bound Execution-State Checkpoint and Restore for Low-Latency, Small-Batch, On-Device Physical-AI Serving}

\author{Liang Su\\
\small \texttt{7thuniversels@gmail.com}\\[2pt]
\small\url{https://github.com/flashrt-project/FlashRT}}

\date{}

\begin{document}
\maketitle

\begin{abstract}
Mainstream LLM serving reuses prefix work through paged or radix key--value (KV)
caches---vLLM's PagedAttention~\cite{paged2023} and SGLang's
RadixAttention~\cite{sglang2024}---which manage one \emph{positionally
addressable fragment} of execution state (the KV cache) under eager or piecewise
execution. Both use CUDA Graphs~\cite{cudagraphs}, but their graphs deliberately
do \emph{not} bind the KV as a self-contained buffer set: attention reads through
mutable block-table inputs so a single captured graph can gather from arbitrary
physical blocks. That indirection
is the precondition for block reuse, but it also means the captured graph plus its
bound buffers never constitute a self-contained, freezable snapshot of the whole
forward pass.

We present a two-layer answer for the \emph{opposite} regime---extreme low latency at
small batch (single- or few-stream), on-device/edge, the physical-world interactive
loops of \S\ref{sec:intro}. \emph{First}, FlashRT is a white-box, backend-facing kernel
runtime (this paper evaluates its NVIDIA CUDA backend) whose hot path is a captured
\emph{graph} over \emph{contiguous static}
buffers with no block-table indirection (in the measured LLM path a single CUDA graph
launch in the NVIDIA backend for the selected shape bucket; the contract equally permits a
DAG of named subgraphs---granularity is a performance choice, not the abstraction). On
the identical hybrid model and GPU its cold time-to-first-token is
$2.6$--$2.8\times$ lower than vLLM's, a measured low-latency \emph{execution floor} for
this setup. \emph{Second}, because that computation graph runs over a fixed, named
buffer set, the complete state needed to continue the next replay at any committed
token boundary is exactly that closed, named buffer set; freezing it---the
\emph{execution-state capsule}---is a
\textbf{checkpoint and restore of the graph-bound execution state}, turning
\code{restore}, \code{fork}, and \code{rollback} of a session into a single copy of
that buffer set, and prefix reuse from a recompute problem (compute bound) into a
bandwidth-bound copy-and-rebuild that \emph{preserves} the floor across session reuse,
branch, interruption, and re-entry. (We use ``graph-bound execution state'' rather than
``whole graph'': the captured plan may be one graph or several subgraphs---what is
checkpointed is the closed buffer set its live state depends on.) Crucially these are
not two independent features: the same design choice---static contiguous buffers and
graph replay over them, no block-table indirection---is simultaneously what makes the
runtime fast and what makes the state freezable. The capsule rests on a minimal execution contract---three
handle types, one opaque shape key, a strict mechanism-not-policy
boundary---which it extends by a single mechanism. One minimal contract spans three
scenarios: \emph{at the execution-mechanism level}, an LLM coding agent's warm start
and a robot reinforcement-learning (RL) rollout's episode reset are the same
snapshot/restore verb, and a hierarchical planner--actor hand-off is the contract's
zero-copy buffer pass.

On an RTX 5090, capsule restore is exact at the tested level: byte-identical stored
state, token-identical greedy decode for the LLM, and byte-identical action replay
(reported also as cosine $1.0$) for a vision-language-action (VLA) diffusion policy---including a
chunk-alignment condition that exact reuse of a chunked linear-attention scan
requires. GPU-resident snapshot and restore are sub-millisecond (host/disk tiers add
a one-time transfer), and the time-to-first-token speedup over a cold prefill
\emph{widens with prefix length} ($3.9\times$ at $2$k to $27\times$ at $16$k; measured
to the first base-logit token, with decode-side MTP excluded for fairness). Measured
to that same TTFT, the capsule is \emph{lower} than vLLM's automatic prefix caching on
the warm reuse it does ($\sim$$1.4$--$2.8\times$), while
additionally reusing the recurrent state that block/radix caches do not expose as a
first-class managed object---an ablation confirms the gain is the state mechanism,
not the runtime alone (a restore that keeps the positional KV but drops the
recurrent fold diverges, while the full capsule is token-exact). In one line, capsules
move the unit of reuse \emph{from token-addressed KV fragments to graph-bound
execution-state boundaries}: this is not a better KV cache, but a latency-first runtime
substrate plus a third managed object---execution state---that together define a
serving design point high-throughput stacks intentionally do not optimize for. We
replicate the LLM results and robot-policy mechanism tests on two
unified-memory on-device systems---a Jetson AGX Thor (\code{sm\_110}) and a DGX Spark (GB10,
\code{sm\_121})---where the same correctness and structural properties hold; on Thor,
whose cold prefill costs seconds, the cold$\to$capsule speedup widens to $9$--$76\times$.
\end{abstract}

\section{Introduction}
\label{sec:intro}

\paragraph{Premise and positioning.} vLLM~\cite{paged2023} and SGLang~\cite{sglang2024}
are the state-of-the-art serving infrastructures; each defines a clear, highly
efficient regime---maximizing aggregate throughput under high concurrency, with
KV-cache management (paging, radix prefix reuse) engineered to that end---and we
take that as given, not as something to beat. By \emph{physical-AI serving} we mean
low-latency, small-batch (single- or few-stream) interactive inference loops whose
outputs drive language, speech, or action in real time---latency-first LLMs (coding
agents/assistants), voice/TTS front-ends, and vision-language-action (VLA)/robot
policies---typically on one on-device or edge GPU. Our observation is that under the
constraints of \emph{single-stream, on-device, physical-AI serving}---one or a few
interactive streams, a small on-device VRAM
budget, intermittent sessions bound to a control loop, and hard responsiveness
deadlines---those throughput-optimized mechanisms cannot reach their peak
efficiency, because the assumption that makes them efficient (many concurrent
requests amortizing a shared, in-process, automatically-managed cache) does not
hold. The issue is not kernel quality or implementation maturity; it is that the
managed object (a positional KV cache) and its automatic retention policy are
optimized for a different objective. Concretely, the limiting question in this regime
is not only whether a shared prefix can be \emph{matched} and its KV reused---paged
and radix caches do that well, and the KV cache remains essential---but whether the
system can quickly recover a \emph{valid continuation state} after the interaction
changes: a new turn or branch, an interrupt, a re-entry, a fresh session bound to a
control loop. A positional KV cache is part of that state, not the whole control
surface for it. This paper builds the serving system for \emph{that} regime, and the system has
\emph{two} layers, because the regime needs two things: a low-overhead
\emph{execution substrate} that makes a single stream's cold path cheap, and
\emph{explicit control of execution state} that keeps it cheap across reuse. FlashRT%
\footnote{Name note. This work is unrelated to the 2026 arXiv work ``FlashRT'' on
efficient red-teaming for prompt injection and knowledge corruption; here FlashRT
denotes a kernel-level serving/runtime system and its execution-state capsule mechanism.}
provides both---a latency-first white-box runtime (\S\ref{sec:substrate}) and, on top
of it, the execution-state capsule (\S\ref{sec:capsule}). The runtime establishes the
low-latency execution floor; the capsule preserves that floor across session reuse,
branch, interruption, and re-entry. It is a complementary design point, not a replacement.

\paragraph{What ``latency-first'' means here (and what it does not claim).} We use
\emph{latency-first / single-stream lowest latency} in a precise sense: minimizing
per-request wall-clock latency at concurrency~$1$, under a fixed model, precision,
hardware, and correctness target, counting graph-replay and state-preparation cost
but excluding tokenizer and network. ``Lowest'' means lowest \emph{observed among the
open serving paths we test under this setup}, not a theoretical optimum. We
correspondingly do \emph{not} claim high-concurrency throughput, distributed/cluster
serving, dynamic arbitrary-shape batching, or cross-node KV reuse; every comparison
in this paper is single-stream (concurrency~$1$) latency.

A serving system is defined by \emph{what it treats as the managed state object}.
PagedAttention~\cite{paged2023} treats serving as memory management: the KV cache
is paged, a block table maps logical positions to arbitrary physical blocks, and
near-zero fragmentation lets the engine batch more requests.
RadixAttention~\cite{sglang2024} treats it as prefix reuse: KV prefixes form a
radix tree so shared prefixes across calls are matched and reused automatically.
Both manage the same object---the KV cache, a positionally addressable
\emph{fragment} of the model's execution state.

\emph{The unit of reuse is the central distinction.} In paged or radix serving,
\emph{tokens} identify the reusable object: a block table maps token positions to KV
pages, and a radix tree matches token prefixes to cached KV subtrees. A capsule keeps
token boundaries only as \emph{metadata}---a position and a prefix digest used for
validation and chunk alignment (\S\ref{sec:correctness})---and the object it reuses is
instead the \emph{committed execution state} at that boundary: the closed set of
graph-bound buffers needed to continue computation, recurrent and convolution state
included. Capsules thus move the unit of reuse \emph{from token-addressed KV fragments
to graph-bound execution-state boundaries}.

This paper introduces a third managed object. FlashRT captures the \emph{entire}
forward pass as a graph plan over contiguous static buffers (with no block-table
indirection); the complete execution state at a committed token boundary is therefore
a fixed set of named device buffers. We freeze that set into an
\emph{execution-state capsule}. We say \emph{graph-bound execution state} (rather than
\emph{whole graph}) for this \emph{closed live-buffer set},
never a requirement that a deployment be captured as one monolithic CUDA
graph: FlashRT may run a single graph or a DAG of named subgraphs (the contract
supports both, \S\ref{sec:contract}), and a capsule snapshots the complete live-buffer
closure at the boundary regardless---graph granularity is a performance choice. All
serving verbs---warm start, fork, branch, episode reset, interruption and
re-entry---become operations on capsules: \emph{state}, rather than memory or
prefixes, is the first-class object (Table~\ref{tab:axes}). The clinching evidence is
structural, not a speed claim: for the hybrid model, restoring positional KV alone is
\emph{incorrect}---the linear-attention recurrent fold is load-bearing---so a full
capsule restore is token-exact while a KV-only restore diverges at the first token
(\S\ref{sec:eval-mech}).

\paragraph{Regime (the premise of this paper).} We do not propose a replacement
for high-throughput serving. We define and build the serving mechanism for the
\emph{opposite} regime: \textbf{single-stream / few-stream, latency-first,
on-device interactive serving for physical AI}---a coding agent, a robot policy,
an interactive assistant running on one consumer or edge GPU at concurrency~$1$.
The dominant cost here is not steady-state throughput but \emph{responsiveness}:
warm-start time-to-first-token (TTFT) for LLMs, time-to-first-action / first-audio
(TTFA) for embodied and streaming models. A coding agent resends the same large
prefix every turn---system prompt, tool schemas, repository index, project memory,
often $10$k--$50$k tokens---and every fresh session or branch cold-prefills it; a
robot must react and re-enter within a control tick. This is the regime today's
throughput-first stacks were not built for. Paged/radix caches own
high-concurrency throughput; capsules own single-stream latency and the embodied,
hybrid, cross-domain cases that come with it. The two are complementary points in
the design space, not competitors---and so \textbf{every comparison in this paper
is single-stream (concurrency~$1$) latency; we never make a throughput claim.}

\begin{table}[t]
\centering
\caption{Three managed objects for serving. The key difference is not whether the
system uses CUDA Graphs (all do), but whether the captured graph plus its bound
buffers \emph{are} a freezable, self-contained execution state.}
\label{tab:axes}
\small
\begin{tabular}{@{}llll@{}}
\toprule
System & Managed object & Addressed by & Enables \\
\midrule
vLLM (PagedAttention) & KV memory pages & position (block) & high-throughput batching \\
SGLang (RadixAttention) & KV prefix subtree & prefix (radix) & cross-call prefix reuse \\
\textbf{FlashRT (Capsule)} & \textbf{graph-bound execution state} & \textbf{boundary} & \textbf{restore / fork / rollback} \\
\bottomrule
\end{tabular}
\end{table}

\begin{table}[t]
\centering
\caption{The two layers, read down the stack. vLLM/SGLang optimize the top of the
stack (substrate) for throughput and the middle (managed object) for automatic KV
reuse; FlashRT optimizes the substrate for single-stream latency and makes the
managed object the whole execution boundary under explicit policy. The capsule
(middle/bottom rows) only makes sense \emph{because} of the substrate row.}
\label{tab:layers}
\small
\begin{tabular}{@{}lll@{}}
\toprule
Layer & vLLM / SGLang & FlashRT \\
\midrule
execution substrate  & high-throughput batching, flexible KV paging & latency-first single-stream execution \\
managed state object & KV blocks / radix prefixes & whole execution boundary \\
reuse control        & automatic cache policy (LRU) & explicit serving policy (pin/fork) \\
\bottomrule
\end{tabular}
\end{table}

\paragraph{Contributions.}
\begin{enumerate}[leftmargin=1.4em,itemsep=2pt]
\item \textbf{A latency-first single-stream runtime substrate}
(\S\ref{sec:substrate}): a white-box, backend-facing kernel runtime (evaluated here on
its NVIDIA CUDA backend) whose hot path is a captured
\emph{graph plan} over contiguous static buffers with no block-table indirection (a
single backend graph replay---one CUDA graph launch on NVIDIA---in the measured LLM
path, or a DAG of named subgraphs), designed for concurrency-$1$ on-device serving. We show it is a \emph{measured}
low-latency execution floor under this setup: on the identical hybrid model and GPU
its cold TTFT is $2.6$--$2.8\times$ below vLLM's, with a tight latency tail
(\S\ref{sec:eval}). We measure responsiveness (TTFT/TTFA) throughout and treat decode
throughput and speculative decoding as out of scope.
\item \textbf{The execution-state capsule} (\S\ref{sec:capsule}): on that substrate,
the complete restorable state at a committed boundary, frozen as a fixed set of named
device buffers, with four verbs (snapshot, restore, fork, rollback) and a cost
model that turns prefix reuse from recompute into copy-and-rebuild. The same design
choice that makes the substrate fast is what makes the state freezable.
\item \textbf{A correctness envelope} (\S\ref{sec:correctness}): three layers (byte
restore, state completeness, token equivalence); the \emph{stored state} is
byte-exact and the tested paths are \emph{token-identical} under greedy decode (the
VLA action replay is byte-identical, cosine $1.0$)---we do not separately claim
byte-identical logits. A KV-only restore (positional KV without the recurrent fold)
diverges, and we identify the chunk-alignment condition under which exact reuse of a
chunked linear-attention recurrent scan holds.
\item \textbf{Regime evidence across the layers} (\S\ref{sec:domains},
\S\ref{sec:eval}): the runtime floor, the capsule's gain over our own cold path, and
the retention-control gain over an automatic prefix cache under a target embodied
working set---plus one minimal contract serving an LLM coding agent's warm start, a
robot RL rollout's episode reset, and a planner--actor hand-off.
\end{enumerate}

We scope this paper deliberately thin: it establishes the mechanism, its
correctness, and a controlled latency benchmark. Production multi-turn agent
serving and on-robot evaluation are explicitly future work (\S\ref{sec:scope}).

\FloatBarrier
\section{The Latency-First Runtime Substrate}
\label{sec:substrate}
\label{sec:bg}

This is the first layer of the system and the precondition for everything after it:
before the capsule manages \emph{which} state to compute from, the runtime must make
a single stream's computation cheap. We first contrast the throughput-first
substrate that paged/radix engines are built on with the latency-first substrate we
need, then describe FlashRT's design and the one choice that ties the runtime's speed
to the capsule's existence. The substrate is evaluated directly in
\S\ref{sec:eval} (the runtime floor: cold TTFT and tail), not assumed.

\paragraph{The shared premise.} Paged and radix KV caches rest on a common
premise: the KV cache is a positionally addressable set of blocks, and the
attention kernel can gather KV from arbitrary physical blocks each step. This is
what lets vLLM share blocks copy-on-write and lets SGLang match the longest common
prefix in a radix tree. Both systems do use CUDA Graphs for low launch overhead.
In vLLM v1 the default mode runs attention eagerly (``piecewise'') while capturing
the rest, or captures full decode-step graphs; in SGLang the decode step is always
captured and variable-length prefill uses piecewise capture with attention split
out. In every case the attention operator reads KV through a \emph{mutable
block-table / index input} so that one captured graph can be replayed against
different physical blocks on successive steps.

\paragraph{Why this mechanism is mismatched to our regime.} Two constraints make a
block/radix design ill-suited here---by construction, not by preference.

\emph{(1) The captured graph is not a freezable state.} Because attention reads KV
through mutable indices into a separately managed paged pool, the graph plus its
bound buffers never hold the whole forward's state; the real KV lives in the
allocator, addressed by indices that change every step. There is no fixed buffer
set to freeze, fork, or rewind. FlashRT instead captures the whole forward
(attention included) over contiguous static buffers with no indirection---so the
bound buffers \emph{are} the complete state.

\emph{(2) Hybrid recurrent state is not prefix-addressable.} Our flagship LLM is a
hybrid linear-attention / full-attention model. The full-attention KV is
positional and \emph{can} be sliced by prefix, but the linear-attention recurrent
state and convolution state are a \emph{fold over the entire
prefix}~\cite{mamba2023,gateddeltanet2024}: the state at position $N$ is a function
of all $N$ tokens, with no sub-slice for ``the first $1000$ tokens.'' Positional KV
reuse alone cannot reconstruct this state; a radix tree of KV blocks does not expose
it as a reusable object. Snapshotting the whole state does.

The viewpoint flip is: others give up self-contained graph state to gain block
flexibility; FlashRT keeps a closed static-buffer graph plan and buys prefix reuse
back with a state snapshot. This is the crux of the differentiation, and it is sharp.
Reusing a \emph{shared text prefix} is not what distinguishes capsules---vLLM's
automatic prefix caching and SGLang's radix tree already do that, and we make no claim
to beat them on it. What a block/radix cache does not expose as a \emph{first-class
managed object}, and a capsule does, is three things: (i)~reuse of \emph{hybrid
recurrent state}, which has no addressable block; (ii)~\emph{fork} of one whole
boundary into $N$ live sessions; and (iii)~\emph{rollback} to an earlier boundary.

\emph{To be precise about the claim:} we do not argue that paged/radix engines
\emph{cannot} be extended to copy extra state---any system can add buffers. The
distinction is whether the \emph{whole continuation state is the first-class managed
object}. In paged/radix systems it is not: the graph is replayable but not
self-contained (attention reads KV through mutable indices into an external pool, and
recurrent/conv state is managed in a separate cache), so a whole-boundary
snapshot/restore/fork would mean adding a \emph{new} state-snapshot object
\emph{outside} the KV/radix cache. That added object is precisely what we call a
capsule---which FlashRT can make first-class because it already holds the whole
committed state as a closed buffer set, not addressable fragments, while keeping the
latency of static-buffer graph-plan replay.

\paragraph{The substrate: a white-box latency-first runtime.} FlashRT is a
white-box, backend-facing kernel runtime with a thin per-model pipeline, not a compiler or
graph rewriter, and it is built for concurrency-$1$ latency rather than aggregate
throughput; the backend evaluated in this paper is NVIDIA CUDA. Memory-bound operators
(normalization, activations, fused
residual/norm/quant, QKV split with rotary embedding) are hand-written backend kernels over
contiguous buffers (CUDA kernels in the NVIDIA backend); GEMMs are thin wrappers over the
backend's vendor libraries (cuBLASLt / CUTLASS FP8 and NVFP4 on NVIDIA).
Setup work (weight load, calibration, autotuning, capturing the whole forward as a
graph plan) runs once in a Python frontend; in the measured LLM path the hot path is
a single backend graph replay (one \code{cudaGraphLaunch} on NVIDIA) that is byte-identical to the un-captured baseline,
while the contract equally permits a DAG of named subgraphs (\S\ref{sec:contract})---
the load-bearing property is the closed buffer set, not the graph count. This is
the layer we evaluate as the \emph{runtime floor} (\S\ref{sec:eval}): on the same
hybrid model and GPU it has $2.6$--$2.8\times$ lower cold TTFT than vLLM with a tight
latency tail. We measure responsiveness (TTFT/TTFA) and treat steady-state decode
throughput and speculative decoding as orthogonal and out of scope.

\paragraph{One design choice, two consequences.} The decisive point is that this is
\emph{not} two separate features bolted together. vLLM/SGLang accept block-table
indirection to win block flexibility (high throughput, many requests, low
fragmentation), at the cost that the graph-bound state is never self-contained.
FlashRT makes the opposite choice---static contiguous buffers and graph-plan replay
over them, no indirection---and that single choice is simultaneously \emph{why the kernels are
fast} (no gather, byte-identical replay, no per-step launch/Python overhead)
\emph{and why the forward can be captured whole} so that the state at any boundary is
a fixed, freezable buffer set. \textbf{Latency-first execution and capsule state
management are two consequences of one design choice, not two independent
mechanisms.} Freezing that buffer set is the capsule (\S\ref{sec:capsule}).

\FloatBarrier
\section{Execution-State Capsules}
\label{sec:capsule}

\begin{figure}[t]
\centering
\begin{tikzpicture}[
  font=\scriptsize, >=Stealth,
  lab/.style={font=\footnotesize\bfseries},
  sub/.style={font=\scriptsize\itshape, text=gray!55!black},
  note/.style={font=\scriptsize\itshape, text=gray!55!black},
  rbox/.style={draw=red!55!black, fill=red!5, rounded corners=2pt, align=center, inner sep=2.5pt},
  rcell/.style={draw=red!55!black, fill=red!10, minimum height=5.2mm, minimum width=7mm, inner sep=0pt},
  gcell/.style={draw=green!45!black, fill=green!12, minimum height=6mm, inner sep=2.5pt, align=center},
  gbox/.style={draw=green!45!black, fill=green!8, rounded corners=2pt, align=center, inner sep=2.5pt},
  ext/.style={draw=gray!55, dashed, rounded corners=2pt, align=center, inner sep=2.5pt},
  rverd/.style={font=\scriptsize, text=red!65!black, align=left},
  gverd/.style={font=\scriptsize, text=green!45!black, align=left},
  arr/.style={->, shorten >=1pt, gray!55!black},
]
\node[lab,anchor=west] at (0,0.4) {(a) Paged / radix (vLLM, SGLang)};
\node[sub,anchor=west] at (0,0.0) {managed object: \emph{token}-indexed KV};
\node[rbox,anchor=west,text width=24mm] (tok) at (0.5,-0.8) {tokens $t_0\dots t_N$};
\node[rbox,anchor=west,text width=24mm] (m) at (0.5,-1.65) {radix match / block table};
\node[rcell,anchor=west] (p1) at (0.5,-2.6) {$B_3$};
\node[rcell,right=0pt of p1] (p2) {$B_0$};
\node[rcell,right=0pt of p2] (p3) {$B_2$};
\node[rcell,right=0pt of p3] (p4) {$B_1$};
\node[note,anchor=west] at (3.6,-2.6) {paged pool (mutable index)};
\node[ext,anchor=north west,text width=46mm] (mm) at (0.5,-3.15)
  {recurrent $+$ conv: a \emph{separate} cache (no block)};
\draw[arr] (tok)--(m);
\draw[arr] (m.south)--(p2.north);
\node[rverd,anchor=north west,text width=64mm] at (0,-3.95)
  {$\Rightarrow$ live state is spread over an indexed pool $+$ side caches: \\ the captured graph is \emph{not} self-contained};
\draw[gray!45,dashed] (7.35,0.6) -- (7.35,-4.6);
\node[lab,anchor=west] at (7.8,0.4) {(b) FlashRT capsule};
\node[sub,anchor=west] at (7.8,0.0) {managed object: the \emph{execution state} at $P$};
\node[sub,anchor=west] at (7.8,-0.62) {one contiguous, graph-bound buffer set:};
\node[gcell,anchor=west,text width=13mm] (kv) at (7.8,-1.35) {KV $[0{:}P)$};
\node[gcell,right=0pt of kv,text width=13mm] (rc) {recurrent};
\node[gcell,right=0pt of rc,text width=7mm] (cv) {conv};
\node[gcell,right=0pt of cv,text width=7mm] (mt) {MTP};
\node[gcell,right=0pt of mt,text width=8mm] (me) {meta};
\node[gbox,anchor=north west,fill=green!16,text width=30mm] (cap) at (8.7,-2.5)
  {\code{snapshot}: freeze the \emph{whole} set $=$ \textbf{capsule}};
\draw[arr,green!50!black,thick] (rc.south|-kv.south)--(cap.north);
\node[gverd,anchor=north west,text width=64mm] at (7.8,-3.95)
  {$\Rightarrow$ the live state \emph{is} one freezable object: \\ no indirection, no side caches};
\end{tikzpicture}
\caption{What each system manages, drawn concretely. \emph{(a)} A paged/radix engine
addresses reuse by \emph{token}: a radix match or block table maps token
positions/prefixes to KV pages scattered in an external pool through \emph{mutable
indices} (shown out of order, $B_3 B_0 B_2 B_1$), and the hybrid recurrent/convolution
state sits in a \emph{separate} cache with no addressable block---so the captured graph
is replayable but its live state is spread across an indexed pool and side caches, not
self-contained. \emph{(b)} FlashRT captures the forward as a graph plan over
\emph{contiguous static} buffers, so the committed boundary's live state \emph{is} one
named, ordered buffer set $\{$KV, recurrent, conv, MTP, meta$\}$ bound to the graph;
freezing that set whole is the capsule. The unit of reuse moves from a token-addressed
KV fragment to the graph-bound execution-state boundary.}
\label{fig:managed-objects}
\end{figure}
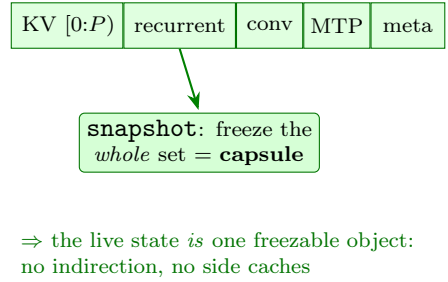

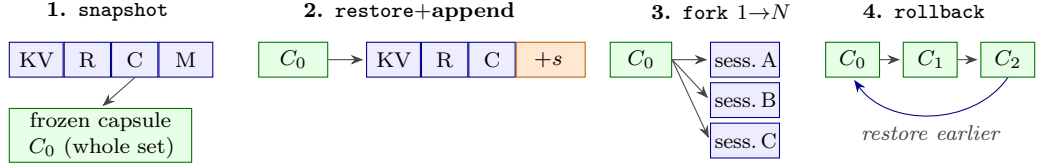
\begin{figure}[t]
\centering
\begin{tikzpicture}[
  font=\scriptsize, >=Stealth,
  pl/.style={font=\scriptsize\bfseries},
  note/.style={font=\scriptsize\itshape, text=gray!50!black},
  live/.style={draw=blue!55!black, fill=blue!7, minimum height=4.6mm, inner sep=1.6pt, align=center},
  froz/.style={draw=green!45!black, fill=green!10, minimum height=4.6mm, inner sep=1.6pt, align=center},
  hot/.style={draw=orange!75!black, fill=orange!18, minimum height=4.6mm, inner sep=1.6pt, align=center},
  arr/.style={->, shorten >=1pt, gray!55!black},
]
\node[pl,anchor=west](t1) at (0,0) {1. \texttt{snapshot}};
\node[live,anchor=north west,below=4mm of t1.west,text width=6mm](l1k){KV};
\node[live,right=0pt of l1k,text width=5mm](l1r){R};
\node[live,right=0pt of l1r,text width=5mm](l1c){C};
\node[live,right=0pt of l1c,text width=6mm](l1m){M};
\node[froz,below=4mm of l1k.south west,anchor=north west,text width=23mm](c0){frozen capsule $C_0$ (whole set)};
\draw[arr](l1c.south)--(c0.north);
\node[pl,anchor=west](t2) at (3.4,0) {2. \texttt{restore}$+$append};
\node[froz,anchor=north west,below=4mm of t2.west,text width=8mm](r0){$C_0$};
\node[live,right=5mm of r0,text width=6mm](l2k){KV};
\node[live,right=0pt of l2k,text width=5mm](l2r){R};
\node[live,right=0pt of l2r,text width=5mm](l2c){C};
\node[hot,right=0pt of l2c,text width=8mm](l2s){$+s$};
\draw[arr](r0)--(l2k.west);
\node[pl,anchor=west](t3) at (8.0,0) {3. \texttt{fork} $1{\to}N$};
\node[froz,anchor=north west,below=4mm of t3.west,text width=7mm](f0){$C_0$};
\node[live,right=5mm of f0,text width=8mm](sa){sess.\,A};
\node[live,below=0.6mm of sa,text width=8mm](sb){sess.\,B};
\node[live,below=0.6mm of sb,text width=8mm](sc){sess.\,C};
\draw[arr](f0.east)--(sa.west);
\draw[arr](f0.east)--(sb.west);
\draw[arr](f0.east)--(sc.west);
\node[pl,anchor=west](t4) at (10.8,0) {4. \texttt{rollback}};
\node[froz,anchor=north west,below=4mm of t4.west,text width=6mm](v0){$C_0$};
\node[froz,right=3mm of v0,text width=6mm](v1){$C_1$};
\node[froz,right=3mm of v1,text width=6mm](v2){$C_2$};
\draw[arr](v0)--(v1); \draw[arr](v1)--(v2);
\draw[arr,blue!55!black] (v2.south)to[out=235,in=305] node[below,note]{restore earlier}(v0.south);
\end{tikzpicture}
\caption{The serving verbs as operations on the buffer set (cf.\ Algorithm~\ref{alg:capsule}).
\textbf{1.~\texttt{snapshot}} freezes the whole live set (KV, recurrent R, conv C, MTP M)
into a tiered capsule $C_0$. \textbf{2.~\texttt{restore}} copies $C_0$ back into the live
buffers and appends only the new suffix $s$, then \emph{replays the already-captured
graph}---no recapture, no prefix recompute. \textbf{3.~\texttt{fork}} restores one
$C_0$ into $N$ independent live sessions (token-exact, \S\ref{sec:eval-thor}).
\textbf{4.~\texttt{rollback}} restores an \emph{earlier} committed boundary of the same
session (undo a turn / retry). A \emph{hard interrupt} is the host overwriting a bound
sub-buffer (e.g.\ a subgoal) between replays---consumed on the next tick with no
recapture (\S\ref{sec:eval}). A KV/radix object exposes none of \texttt{fork}
($1{\to}N$ whole boundary), \texttt{rollback}, or recurrent-state reuse as a first-class
operation (Table~\ref{tab:cap}).}
\label{fig:verbs}
\end{figure}

\paragraph{Definition.} A capsule freezes a \emph{committed} execution boundary at
position $P$ into a fixed set of named device buffers. For the hybrid LLM it
contains: a small fixed-size part (linear-attention recurrent state, convolution
state, the multi-token-prediction (MTP) tail and compact cache with its valid
range, the last hidden as an MTP seed, and boundary metadata such as
\code{cur\_pos} and a token-prefix digest); and the KV region (the persistent
full-attention KV valid over $[0,P)$, plus the long-context FP8 dequantization
stage's valid end). The small part is fixed-size and cheap to snapshot; the KV
region grows with $P$ and dominates the footprint (FP8-KV roughly halves it).
Capsules are therefore taken only at \emph{meaningful} boundaries---a pinned shared
prefix, an episode start, a turn boundary---never on a dense token grid.

\paragraph{The graph executable is not the capsule.} A captured graph alone (a CUDA
Graph in the NVIDIA backend) is only a
\emph{replayable computation}---a command DAG for ``how to compute''---and vLLM and
SGLang capture graphs too. The capsule is that computation \emph{together with} the
exact buffer state at a committed boundary: the byte snapshot of the graph-bound
live buffers (KV, recurrent, conv, MTP, metadata) plus the boundary descriptor.
Restore copies those bytes back into the live buffers and replays the \emph{same}
captured graph. Low latency comes from the two stacked: graph replay avoids launch
/ Python / recapture overhead, and state restore avoids prefix recompute. The new
managed object is therefore the \emph{execution state}, not the graph---which is why
``we also use CUDA Graphs'' is not the claim. What makes it freezable is that
FlashRT's graph is captured over contiguous static buffers, so the boundary state is
a fixed, self-contained buffer set (Table~\ref{tab:axes}); a graph whose attention
reads a mutable paged pool has no such self-contained set to freeze.

\begin{table}[t]
\centering
\caption{Cost structure. Capsule turns shared-prefix reuse from a compute-bound
recompute into a bandwidth-bound state copy. Both scale with $L$; the win is the
slope, not a constant-time restore.}
\label{tab:cost}
\small
\begin{tabular}{@{}lll@{}}
\toprule
Operation & Cost & Scaling \\
\midrule
\code{snapshot} & freeze a small $+$ KV buffer set & $\Theta(\text{bytes})$, bandwidth-bound copy \\
\code{restore} & copy bytes back $+$ re-bind (lazy dequant) & $\Theta(\text{bytes})$, bandwidth-bound copy \\
cold prefill & recompute the shared prefix & compute-bound, $\propto$ prefix length \\
\bottomrule
\end{tabular}
\end{table}

\paragraph{Cost model.} Let $L$ be the shared-prefix length. A cold turn pays
$T_{\text{prefill}}(L+s)$ for prefix plus suffix $s$; a capsule turn pays
$T_{\text{restore}}(L) + T_{\text{append}}(s)$. Both touch $L$, but at very
different rates: cold \emph{recomputes} the prefix (compute-bound, and on the
short route also re-captures a graph per position), whereas restore only
\emph{copies} the capsule's $\Theta(L)$ bytes at memory bandwidth and re-binds a
bounded boundary (the FP8 BF16 working stage is rebuilt \emph{lazily} on the next
read, not eagerly in restore). The benefit is the compute-vs-bandwidth asymmetry:
\[
  T_{\text{prefill}}(L) \;\gg\; T_{\text{restore}}(L),
\]
and since prefill grows with $L$ while a bandwidth copy of the state stays cheap,
the gap widens with $L$. By construction the capsule touches only prefill/TTFT,
never steady-state decode. We do not claim restore is $O(1)$: it is $\Theta(L)$
bytes of copy: the point is that copying state is far cheaper than recomputing it.

\paragraph{Four verbs.} (i)~\emph{snapshot}: freeze a boundary (to GPU, host RAM,
or disk). (ii)~\emph{restore}: copy the capsule back into the live buffers and
rebuild the boundary, then reuse the \emph{same captured graphs}---no recapture.
(iii)~\emph{fork}: restore one capsule into several sessions---one prefill of a
shared prefix feeds $N$ branches (tree-of-thought, best-of-$N$, parallel
tool-calls). (iv)~\emph{rollback}: restore an \emph{earlier committed boundary} of the
same session (undo a turn / retry from a checkpoint). Verbs (iii) and (iv) are not
native to a KV/radix cache alone; supporting them as whole-session operations would
require an additional state-snapshot object (the capsule makes that object explicit).
Algorithm~\ref{alg:capsule} gives the mechanism: each verb is a byte-copy of the
boundary's buffer closure plus a replay of the already-captured graph---never a
recapture or a prefix recompute.

\begin{algorithm}[t]
\DontPrintSemicolon
\SetKwProg{Fn}{Function}{:}{}
\SetKwFunction{snap}{snapshot}\SetKwFunction{rest}{restore}\SetKwFunction{frk}{fork}
\KwData{live buffer set $B=\{$KV,\,recurrent,\,conv,\,MTP$\}$ over contiguous static
buffers; chunk size $C$; captured graph table \code{Graph[key]}}
\Fn{\snap{$P$}}{
  $P' \gets \lfloor P/C\rfloor\cdot C$ \tcp*{chunk-align the boundary (\S\ref{sec:correctness})}
  $cap.\text{bytes} \gets$ device-copy of $B$ valid at $P'$\;
  $cap.\text{desc} \gets (P',\ \text{token-prefix digest},\ \text{KV/dequant valid-ends})$\;
  \Return tiered handle $cap$ (GPU $\mid$ host $\mid$ disk)\;
}
\Fn{\rest{$cap$, $suffix$, $key$}}{
  \textbf{assert} $cap.\text{desc.digest}$ matches the deployment (weights, quant, kernel, bucket)\;
  \lIf{$cap$ not GPU-resident}{promote $cap$ \tcp*[f]{only non-sub-ms step}}
  copy $cap.\text{bytes}$ back into $B$; rebind $cap.\text{desc}$\;
  replay \code{Graph[$key$]}; append $suffix$ \tcp*{no recapture, no prefix recompute}
}
\Fn{\frk{$cap$, $n$}}{
  \Return $n$ sessions, each \rest{$cap,\cdot,\cdot$} \tcp*{one boundary $\to N$ live sessions}
}
\caption{Capsule lifecycle over the committed boundary's buffer closure.
\code{rollback} is \code{restore} of an \emph{earlier} boundary of the same session.}
\label{alg:capsule}
\end{algorithm}

\FloatBarrier
\section{The Execution Contract}
\label{sec:contract}

Capsules sit on a minimal C ABI with zero dependency on the kernel layer
(Listing~\ref{lst:abi}). It fixes \emph{mechanism}, never scenario \emph{policy}.

\begin{lstlisting}[caption={Core of the execution contract (abridged from the
188-line header). It sees only streams, graphs, events, and named buffers.},
label={lst:abi},captionpos=b]
/* Buffer: the only state primitive. KV, scales, noise, subgoal
   embeddings are all Buffers; the framework owns lifetime+pointer,
   never append/fork/evict verbs. */
frt_buffer frt_buffer_alloc(frt_ctx, const char* name, size_t bytes);
frt_buffer frt_buffer_wrap (frt_ctx, const char* name, void* dptr, size_t);
int        frt_buffer_copy (frt_ctx, frt_buffer dst, size_t doff,
                            frt_buffer src, size_t soff, size_t n, int stream);

/* Graph: a table ShapeKey -> captured graph-exec. Capture our own, or
   adopt an externally instantiated (e.g. torch) graph-exec. */
int frt_graph_capture(frt_graph, frt_shape_key, void(*rec)(void*,void*), void*);
int frt_graph_adopt  (frt_graph, frt_shape_key, void* external_graph_exec);
int frt_graph_bind   (frt_graph, const char* port, frt_buffer);
int frt_graph_replay (frt_graph, frt_shape_key, int stream_id);

/* Plan: a dumb DAG of (graph,key) nodes; data dependencies only. */
/* ShapeKey: opaque u64 encoding (B,S,...); batch is one field, not an axis. */
\end{lstlisting}

The contract is \emph{capture-agnostic}: a captured graph may wrap FlashRT kernels,
torch ops, or native backend kernels, which is why one contract drives both LLM and VLA models.
Two graphs sharing one \code{Buffer} on matching ports is the entire multi-model
hand-off mechanism (zero copy). The \emph{mechanism-not-policy} rule is a hard
boundary: the contract never learns about sessions, KV append/fork/evict, or
schedulers; those live one layer up (a four-layer stack:
\code{serving/} policy $\to$ \code{flash\_rt/} frontend $\to$ \code{exec/} contract
$\to$ \code{csrc/} kernels). The capsule is the rule's best example: it adds
\emph{one} mechanism to the contract---host-backed buffers plus cross-space
asynchronous copy, so a capsule can be parked off-GPU---and everything else
(digest matching, pinning, LRU, restore-versus-rebuild) is serving-layer policy.

\paragraph{Serving-layer semantics.} The contract is deliberately mechanism-only,
but the capsule surfaces as serving \emph{verbs} one layer up
(Listing~\ref{lst:serve}): a session snapshots a named, tiered, optionally pinned
boundary; restores it with a suffix and shape key; forks it into $N$ branches; and a
registry promotes/evicts capsules across the GPU\,$\to$\,host\,$\to$\,disk tiers
under an explicit policy. These are the operations a serving system exposes; the
contract below them never learns what a ``session'' is.

\begin{lstlisting}[caption={Serving-layer capsule semantics (policy layer, above the
contract). The names denote what the serving system offers; pinning, tiering, and
eviction are explicit policy, not automatic cache behaviour.},
label={lst:serve},captionpos=b,language=Python]
cap = session.snapshot(boundary="turn", tier="gpu", pin=True)
session.restore(cap, suffix=new_tokens, shape_key=key)  # no recapture
branches = session.fork(cap, n)            # 1 boundary -> N live sessions
session.rollback(cap_earlier)              # restore an earlier committed boundary
registry.promote(cap, tier="gpu"); registry.evict(policy="lru")
\end{lstlisting}

\code{restore} is the load-bearing verb; its mechanism is the small, fixed sequence of
Algorithm~\ref{alg:capsule} (digest check, GPU-tier promotion if needed, byte-copy of
the buffer closure, metadata rebind, captured-graph replay---no recapture). Pinning is
the one policy knob: a pinned capsule is exempt from \code{evict} while its session is
live; under GPU pressure the registry demotes \emph{unpinned} capsules to host/disk
first, and a restore of a demoted capsule pays one promotion transfer (the only
non-sub-millisecond step).

\FloatBarrier
\section{Correctness: Byte-State Restore, Token Equivalence, and Chunk Alignment}
\label{sec:correctness}

We define correctness in three layers, of increasing strength. (1)~\emph{Byte
restore}: the stored buffers are copied back byte-for-byte---trivially exact in
isolation. (2)~\emph{State completeness}: the capsule names \emph{every} live buffer
the next step depends on, with nothing dangling. We test this adversarially:
snapshot, fully overwrite every live buffer with an unrelated prompt and decode it,
then restore and decode---the output is token-identical to never having run the
other prompt; a missed recurrent register, convolution window, MTP entry, or
metadata field would diverge. (3)~\emph{End-to-end equivalence}: greedy decode after
restore is \emph{token-identical} to the path it replaces (pure restore vs.\ a cold
prefill of the same prefix; restore$+$append vs.\ the append path; fork branches
match). We verify (1)--(3); we report token-level equivalence under greedy decode
and do not separately claim byte-identical logits/hidden states (only that the
emitted token stream matches).

\paragraph{Recurrent-state completeness: the differentiator.} The capsule snapshots
and restores the linear-attention recurrent state and convolution state, not just
the positional KV. This is the reuse a block/radix cache does not expose as a
first-class object, and we verify it is \emph{exact}: at a chunk-aligned boundary,
restore$+$append reproduces a cold full prefill token-for-token, recurrent state
included (below).

\paragraph{The chunk-alignment condition.} The long chunked linear-attention
prefill folds its recurrent state \emph{per chunk}, so the state at a position
depends on where chunk boundaries fall. A cold full prefill of length $F$ places
boundaries at multiples of the prefill chunk size $C$. If a capsule/append boundary
$P$ is \emph{not} a multiple of $C$, the append introduces a chunk split the cold
prefill never had, and the two diverge under FP8 rounding (small, but it compounds
through greedy decode). The fix respects model structure rather than working
around it: snapshot at an aligned boundary $P' = \lfloor P/C\rfloor\cdot C$. Then
\[
  \text{restore}(P') + \text{append}(\text{suffix}) + \text{decode}
  \;=\;
  \text{cold-full-prefill} + \text{decode}
\]
token-for-token, with the sub-chunk remainder ($<C$ tokens) cheaply re-prefilled by
the append. This is itself a finding: \emph{exact reuse of a chunked recurrent scan
requires respecting its chunk boundaries.}

\FloatBarrier
\section{One Mechanism, Three Domains}
\label{sec:domains}

Because a capsule is just ``the committed state as a set of buffers,'' one
\emph{contract} spans LLM, VLA, and robot control. We are precise about what is
unified: the LLM warm start and the robot episode reset are the \emph{same}
snapshot/restore verb; the planner--actor case needs \emph{no} capsule and uses the
contract's zero-copy hand-off instead. The claim is therefore that one minimal
contract expresses this spectrum---restore along the time axis, hand-off along the
space axis---not that every scenario needs a capsule. Figure~\ref{fig:scenarios}
draws the four scenarios concretely.

\begin{figure}[t]
\centering
\begin{tikzpicture}[
  font=\scriptsize, >=Stealth,
  pl/.style={font=\footnotesize\bfseries},
  note/.style={font=\scriptsize\itshape, text=gray!55!black},
  cps/.style={draw=blue!55!black, fill=blue!9, rounded corners=1pt, align=center, inner sep=2.2pt},
  suf/.style={draw=green!50!black, fill=green!13, rounded corners=1pt, align=center, inner sep=2.2pt},
  oup/.style={draw=brown!65!black, fill=yellow!28, rounded corners=1pt, align=center, inner sep=2.2pt},
  ovr/.style={draw=orange!80!black, fill=orange!22, rounded corners=1pt, align=center, inner sep=2.2pt},
  lgc/.style={minimum size=2.6mm, inner sep=0pt},
  arr/.style={->, shorten >=1pt, gray!55!black},
]
\node[cps,lgc] (lg1) at (0,0.85) {}; \node[anchor=west,right=1mm of lg1] {pinned capsule (invariant state)};
\node[suf,lgc] (lg2) at (5.7,0.85) {}; \node[anchor=west,right=1mm of lg2] {appended suffix / fresh input};
\node[oup,lgc] (lg3) at (10.8,0.85) {}; \node[anchor=west,right=1mm of lg3] {model output};
\node[ovr,lgc] (lg4) at (13.7,0.85) {}; \node[anchor=west,right=1mm of lg4] {overwrite / interrupt};
\node[pl,anchor=west] at (0,0.05) {(a) Coding agent: warm multi-turn};
\node[cps,anchor=west,text width=25mm] (a0) at (0,-0.85) {pinned prefix $C_0$\\system $+$ tools $+$ repo};
\node[suf,anchor=west] (a1) at (3.7,-0.5) {$+$turn 1};
\node[oup,anchor=west,right=2mm of a1] (a1o) {answer 1};
\node[suf,anchor=west] (a2) at (3.7,-1.2) {$+$turn 2};
\node[oup,anchor=west,right=2mm of a2] (a2o) {answer 2};
\draw[arr] (a0.east)|- (a1.west);
\draw[arr] (a0.east)|- (a2.west);
\draw[arr] (a1)--(a1o); \draw[arr] (a2)--(a2o);
\node[note,anchor=north west,text width=36mm] at (0,-1.75) {each turn $=$ \texttt{restore}$(C_0)+$append (not re-prefill)};
\node[pl,anchor=west] at (8.3,0.05) {(b) Fork: tree-of-thought / best-of-$N$};
\node[cps,anchor=west,text width=11mm] (b0) at (8.3,-0.95) {$C_0$\\shared};
\node[suf,anchor=west] (bA) at (10.5,-0.5) {branch A};
\node[suf,anchor=west] (bB) at (10.5,-1.05) {branch B};
\node[suf,anchor=west] (bC) at (10.5,-1.6) {branch C};
\node[oup,anchor=west] (bAo) at (12.7,-0.5) {out};
\node[suf,anchor=west] (bB1) at (12.7,-1.05) {B.1};
\draw[arr] (b0.east)--(bA.west); \draw[arr] (b0.east)--(bB.west); \draw[arr] (b0.east)--(bC.west);
\draw[arr] (bA)--(bAo); \draw[arr] (bB)--(bB1);
\node[note,anchor=north west,text width=37mm] at (8.3,-2.05) {one capsule $\to$ $N$ independent sessions (each token-exact)};
\node[pl,anchor=west] at (0,-3.25) {(c) Robot RL rollout: episode reset};
\node[cps,anchor=west,text width=17mm] (c0) at (0,-4.15) {episode-init $C_0$};
\node[suf,anchor=west] (c1) at (2.9,-4.15) {obs/act chunk};
\node[oup,anchor=west,right=2mm of c1] (c2) {action};
\draw[arr] (c0)--(c1); \draw[arr] (c1)--(c2);
\draw[arr] (c2.south)to[out=270,in=270,looseness=1.6] node[below,note]{per-chunk loop} (c1.south);
\draw[arr,blue!55!black,dashed] (c2.north)to[out=65,in=115,looseness=0.8] node[above,note]{reset $=$ \texttt{restore}$(C_0)$} (c0.north);
\node[pl,anchor=west] at (8.3,-3.25) {(d) Interrupt $+$ re-entry};
\node[cps,anchor=west,text width=15mm] (d0) at (8.3,-4.15) {planner ctx (pinned)};
\node[suf,anchor=west] (dg) at (10.6,-3.8) {subgoal};
\node[oup,anchor=west] (dt1) at (12.6,-3.8) {act $t_1$};
\node[ovr,anchor=west] (dg2) at (10.6,-4.55) {new subgoal};
\node[oup,anchor=west] (dt2) at (12.6,-4.55) {act $t_2$};
\draw[arr] (d0.east)|- (dg.west); \draw[arr] (dg)--(dt1);
\draw[arr] (d0.east)|- (dg2.west); \draw[arr] (dg2)--(dt2);
\draw[arr,orange!80!black] (dg.south)--(dg2.north);
\node[note,anchor=north west,text width=40mm] at (8.3,-5.05) {overwrite the bound subgoal buffer; next replay consumes it --- no recapture};
\end{tikzpicture}
\caption{Physical-AI serving scenarios and the capsule (cf.\ the verbs in
Fig.~\ref{fig:verbs}). \emph{(a)} A coding agent pins the large shared prefix as $C_0$;
every turn \code{restore}s $C_0$ and appends only the new turn, never re-prefilling the
$10$--$50$k-token prefix. \emph{(b)} \code{fork} restores one $C_0$ into $N$ independent
branches (tree-of-thought, best-of-$N$, parallel tool-calls), each continuing
token-exactly. \emph{(c)} A robot RL rollout snapshots the episode-initial boundary;
the host replays one action chunk per tick, and episode \emph{reset} is
\code{restore}$(C_0)$---no model/graph re-warm. \emph{(d)} A planner pins its context
and writes a \emph{bound} subgoal buffer the actor reads each tick; a mid-run interrupt
\emph{overwrites} that buffer (orange) and the next actor tick consumes the new subgoal
with no graph recapture. All four are the same contract: \code{snapshot}/\code{restore}/%
\code{fork} of state along the time axis (a--c) and a zero-copy bound-buffer hand-off
along the space axis (d).}
\label{fig:scenarios}
\end{figure}

\paragraph{LLM coding agent.} The boundary is a large hybrid state, so the frontend
exposes a model-specific \code{snapshot\_capsule()} / \code{restore\_capsule()}. A
shared prefix is cold-prefilled once and pinned; each later turn, fresh session, or
branch restores it and prefills only the new suffix (\S\ref{sec:eval}).

\paragraph{Robot-policy execution mechanism (offline, not an on-robot result).} A
monolithic \code{while True: act(obs)} rollout loop has no episode-boundary
control---a real complaint from RL users who cannot stop to reset the robot. The fix
is a host-side episode state machine over the contract's interruptible, per-chunk
replay: each loop iteration replays one action chunk, the host checks a termination
predicate (a value-function critic, a keyboard event, or a timeout) between chunks,
and
\emph{reset} restores the episode-initial boundary with no recapture. Here the
boundary is light (a diffusion seed; in production also the observation), so the
\emph{same} capsule mechanism is expressed directly through the contract: a capsule
is a \code{frt} Buffer; snapshot/restore is a \code{frt} device-to-device copy; no
model-specific API. Episode reset is the same verb the coding agent uses.

\paragraph{Planner--actor hand-off (a contrast).} A hierarchical
planner$\to$actor loop needs no capsule: it is a zero-copy buffer hand-off. A
low-rate planner and a high-rate actor co-host in one context; the planner writes a
shared subtask Buffer, the actor reads it each tick, and a mid-episode correction
overwrites that Buffer so the next actor replay consumes the new goal with no
recapture. This marks the boundary: a \emph{capsule} restores or forks a whole
session state along the time axis; a \emph{hand-off} passes a value between live
models along the space axis. Both are mechanisms the contract already provides.

\paragraph{Compute-state recovery, not physical reversal.} A capsule in an embodied
setting recovers \emph{computation}, never the world. The physical state---pose,
object positions, pixels---changes at high frequency and is irreversible; the
capsule makes no attempt to roll it back. What it restores is the \emph{invariant
computational substrate} (planner context, task stack, skill phase, warm
graph/runtime), which a disturbance does \emph{not} invalidate; the volatile
observation is always re-bound from the current world, never from the capsule. The
expensive thing to rebuild after an interruption is not the motion but the
computation---re-prefilling a long planner context and re-warming the runtime---and
that is exactly what the capsule makes cheap.

Restoring stale computation against a changed world would be unsafe, so restore is
gated by \emph{bounded re-entry}: choose the recovery depth by a validity check
(actor state only if the scene is essentially unchanged; otherwise the planner
context with a fresh observation; otherwise a safe fallback), re-observe, verify
preconditions, then resume, replan, or hand to a human. The capsule provides the
mechanism (a fast, byte-exact restore of the stored computational state); the validity predicate and
fallback are serving-layer safety policy. We verify only the mechanism here
(byte-exact stored state, token/action-identical output, \S\ref{sec:eval}); a safety study and on-robot evaluation are
future work (\S\ref{sec:scope}).

\paragraph{When the latency floor holds.} The capsule \emph{recovers} the
restore-not-recompute latency floor---when the selected boundary is valid and
GPU-resident---under exactly this envelope, which covers the common physical-world
disturbances: (i)~\emph{same deployment}---identical weights,
quantization/scales, kernel version, and captured-graph bucket/ShapeKey; (ii)~the chosen
boundary is \emph{pinned in the GPU-resident tier} (host/disk tiers add a one-time
transfer on promotion); (iii)~the suffix falls within a captured shape bucket;
(iv)~for hybrid models, the boundary is \emph{chunk-aligned} (\S\ref{sec:correctness});
and (v)~for embodied control, the volatile observation is re-bound from the current
world and a validity check passes for the chosen depth (actor/skill/planner). A
changing \emph{instruction, subgoal, or interrupt} does not break the floor: the
invariant substrate (planner context, skill, warm runtime) stays pinned and is
restored, while only the new observation/suffix is computed. The floor does
\emph{not} apply when these fail---a new shape needing capture, a host/disk first
restore, changed weights/quant/kernels, an unaligned hybrid boundary, or an
actor-level state invalidated by the world (which falls back to planner re-entry).
This is the precise envelope in which we claim stable low latency.

\FloatBarrier
\section{Evaluation}
\label{sec:eval}

\paragraph{Setup.} The main measurements are on a single NVIDIA GeForce RTX 5090
(\code{sm\_120}, $33.7$\,GB), CUDA~13, and the LLM/robot results are replicated
on-device on a Jetson AGX Thor (\code{sm\_110}) and a DGX Spark (GB10, \code{sm\_121};
both in \S\ref{sec:eval-thor}),
\textbf{single-stream (concurrency $1$)}
throughout---the regime of \S\ref{sec:intro}, not a throughput setting---with the
hybrid LLM in NVFP4 and speculative decode via multi-token-prediction heads
($K{=}3$)~\cite{specdecode2023,medusa2024}. We compare two ways to serve each turn:
\emph{cold} re-prefills prefix$+$suffix every turn; \emph{capsule} prefills the
prefix once, snapshots, then per turn restores and appends only the suffix.
Correctness is checked inline (cold and capsule must emit identical tokens). TTFT is
wall-clock to the first base-logit token; the MTP draft-cache tail fill (decode-side
speculation prep, which the prefill call otherwise runs inline) is excluded, the same
convention on all devices (\S\ref{sec:eval-floor})---so the metric is comparable
across hardware and fair against vLLM, which has no MTP. \emph{Two roles for the baselines, stated up front:} vLLM
is our same-hybrid-model latency baseline (\S\ref{sec:eval-retention}); SGLang appears
structurally, as the radix-prefix managed-object representative (Table~\ref{tab:cap},
from source), and experimentally only through a SGLang-native Higgs-TTS runtime sanity
check (\S\ref{sec:eval-cross}). We bind \emph{performance} numbers to vLLM and
\emph{managed-object} comparison to SGLang. A same-hybrid-model SGLang \emph{latency}
number is deferred for fairness: it is not core to our claims (the
managed-object distinction is established from source), the regime here is single-GPU
single-stream, and SGLang's current release does not support this checkpoint's NVFP4
weight-only quantization---so a clean number is not available without re-quantizing the
model or patching SGLang, both of which would be unfair. We document the full attempt
and this caveat in Appendix~\ref{app:sglang} rather than report a forced number. Peak GPU memory is torch's \code{max\_memory\_allocated}
after model load, graph capture, and all resident state/capsules; for vLLM we use its
reported KV-cache size and reserved GPU memory. Reproduction commands and raw
artifacts are released with the code~\cite{flashrt}.

\paragraph{Target workload.} The regime of \S\ref{sec:intro} is a concrete workload,
and we state it so each experiment maps to a property of it
(Table~\ref{tab:workload}): concurrency $1$--few; a repeated large \emph{stable}
prefix (system prompt, tool schemas, repo/persona context); intermittent
branch/restart and interrupt/re-entry; a working set of skill/subgoal/persona
contexts cycled over time; a hard TTFT/TTFA budget; a limited on-device VRAM budget;
a bounded shape set; correctness requiring exact (or near-exact) state reuse; and
throughput as a non-goal. Our evaluation is organized in three layers---\emph{runtime
floor} (the substrate, \S\ref{sec:substrate}), \emph{mechanism gain} (capsule over
our own cold path), and \emph{retention-control gain} (capsule over an automatic
prefix cache under this working set)---so that the capsule's benefit is never
conflated with the runtime simply being fast.

\begin{table}[h]
\centering
\caption{Target workload properties and the experiment that exercises each.}
\label{tab:workload}
\footnotesize
\begin{tabular}{@{}ll@{}}
\toprule
workload property & experiment \\
\midrule
single-stream runtime floor & cold TTFT p50/90/99 + peak mem (Tab.~\ref{tab:floor}) \\
repeated stable prefix & cold-vs-capsule scaling (Tab.~\ref{tab:break}, Fig.~\ref{fig:scaling}) \\
exact state reuse (hybrid) & KV-only vs full-state ablation (Tab.~\ref{tab:kvonly}) \\
working set + residency pressure & embodied loop, \code{num\_cached\_tokens} (Fig.~\ref{fig:workingset}) \\
interruption / re-entry & LLM+TTS barge-in, composed (Fig.~\ref{fig:bargein}) \\
cross-model hand-off & robot planner$\to$actor + interrupt (\S\ref{sec:eval}) \\
streaming TTFA & Higgs TTS runtime check (Fig.~\ref{fig:higgs}) \\
\bottomrule
\end{tabular}
\end{table}

\begin{table}[h]
\centering
\caption{Evidence level per claim (single-stream; no high-concurrency claim).}
\label{tab:evidence}
\footnotesize
\setlength{\tabcolsep}{4pt}
\begin{tabular}{@{}p{0.27\linewidth}p{0.42\linewidth}p{0.25\linewidth}@{}}
\toprule
claim & evidence & not claimed \\
\midrule
runtime floor (substrate) & cold TTFT p50/90/99 + peak mem (Tab.~\ref{tab:floor}) & decode throughput / MTP \\
LLM capsule reuse & measured TTFT + token-exact (R1, Tab.~\ref{tab:break}) & high-concurrency throughput \\
whole-state (not KV memcpy) & KV-only restore diverges vs full (Tab.~\ref{tab:kvonly}) & a tuned hybrid baseline \\
APC miss mode & embodied working set + \code{num\_cached\_tokens} logs & universal APC failure \\
LLM+TTS barge-in & \emph{composed}: measured LLM re-entry $+$ separate TTS term & co-resident end-to-end pipeline \\
recurrent reuse & token-exact chunk-aligned (R1) + source & a tuned hybrid baseline \\
robot reset / hand-off & offline graph mechanism test, cos$=1.0$ & on-robot task success / safety \\
TTS runtime & single-stream latency sanity check & capsule reuse \\
\bottomrule
\end{tabular}
\end{table}

\paragraph{What we claim, and what we do not.} The baseline is FlashRT's own
no-reuse \emph{cold} path, which isolates the capsule mechanism (reuse vs.\ no
reuse) within one runtime. We do \emph{not} claim to beat vLLM's automatic prefix
caching or SGLang's radix tree on shared-prefix reuse, nor to win on
high-concurrency throughput---those are a different design point
(\S\ref{sec:bg},~\S\ref{sec:scope}). The speedup magnitude below should be read as
``what reuse buys over recompute in this runtime,'' and the differentiating claims
are the structural ones a block/radix cache does not expose as first-class objects:
recurrent-state reuse (\S\ref{sec:correctness}), fork, and rollback, with static-buffer graph capture
preserved.

\paragraph{Correctness gate.} A pytest suite of nine tests is the correctness
contract, all passing on the hardware above: pure-restore equals cold prefill;
restore survives a dirtied state; restore$+$append equals the non-capsule append
path; fork branches match; the chunk-aligned long boundary equals a cold full
prefill; and the not-yet-wired long ``TQ'' KV mode raises rather than producing a
partial capsule. Every capsule result reported below is \emph{token-exact} versus
its cold reference. \emph{Fork and rollback are verified token-exact, not asserted}
(on Thor, \S\ref{sec:eval-thor}): from one capsule, branch~$A$ equals a cold
prefill of \mbox{prefix$+A$} ($0/40$ mismatch) and branch~$B$ equals cold
\mbox{prefix$+B$} with $A\neq B$ (true independent divergence, not identical replay);
and rolling back---descend $A$, restore the earlier boundary, descend $B$---yields
$B=$ cold \mbox{prefix$+B$}. This is exactly the physical-world re-entry invariant for
the LLM: restore the invariant prefix, supply a \emph{changed} input, get a valid
\emph{different} continuation equal to recomputing it.

\FloatBarrier
\subsection{Layer 1: the runtime floor (substrate)}
\label{sec:eval-floor}

Before any reuse, the substrate must make a single stream cheap, and the metric is
\emph{time-to-first-token} (TTFT)---the responsiveness the regime is defined by, not
throughput. \textbf{TTFT convention (used throughout, all devices):} TTFT is the time
to the first \emph{base-logit} token; the MTP draft-cache tail fill, which the prefill
call otherwise runs inline, is decode-side speculation prep (out of scope, \S\ref{sec:capsule})
and is excluded---so the measured TTFT is comparable across hardware and fair against
vLLM (no MTP). We measure the runtime floor on the same hybrid model and GPU (long
FP8-KV route, $4096$-token prefix, $30$ repeats; Table~\ref{tab:floor}). Cold TTFT is
$366.8$\,ms with a \emph{tight tail} (p99 $367.2$\,ms, p90$-$p50 $<0.5$\,ms)---the
hard-responsiveness property the regime needs is met not just at the median but at
the tail---and peak GPU memory is $22.8$\,GB after load and graph capture, so
the low latency is not bought with an outsized memory budget. Against a
throughput-first runtime this same-model cold floor is $2.6$--$2.8\times$ lower
(Table~\ref{tab:vllm}); the capsule (next layer) preserves it at $\sim$$53$\,ms.

\begin{table}[h]
\centering
\caption{Runtime floor, single-stream, $4096$-token prefix (same model/GPU; paper
TTFT, MTP tail excluded). TTFT has a tight tail (p99 within $\sim$$0.5$\,ms of p50) and
peak memory is modest; against a throughput-first runtime the cold floor is
$2.6$--$2.8\times$ lower (Table~\ref{tab:vllm}).}
\label{tab:floor}
\small
\begin{tabular}{@{}lcccc@{}}
\toprule
metric & p50 & p90 & p99 & note \\
\midrule
cold TTFT (ms)        & $366.8$ & $367.2$ & $367.2$ & full prefix$+$suffix prefill \\
capsule TTFT (ms)     & $\phantom{0}53.0$ & $\phantom{0}53.0$ & $\phantom{0}53.1$ & restore$+$append (same run) \\
peak GPU mem (GB)     & \multicolumn{3}{c}{$22.8$ (load $+$ capture)} & capsule $224$\,MB \\
\bottomrule
\end{tabular}
\end{table}

\noindent The tight tail is not specific to $4$k: the per-size breakdown
(Table~\ref{tab:break}) shows capsule TTFT flat at $51$--$57$\,ms and cold rising
$200\to1541$\,ms across $2$k--$16$k, all token-exact, so the hard-responsiveness
property holds across the prefix range, not at one point.

\paragraph{Scope: TTFT/TTFA, not decode throughput.} We deliberately make the
responsiveness metric (TTFT, and time-to-first-audio for streaming) the axis of
every comparison, and we treat \emph{steady-state decode throughput and speculative
decoding (multi-token prediction, MTP) as out of scope}. They are an orthogonal
accelerator of the post-first-token stream: FlashRT uses MTP and this vLLM run does
not, so a head-to-head decode rate would compare two different speculation choices,
not the runtime floor or the capsule mechanism. The first token comes from the base
logit, so MTP does not affect TTFT; the capsule touches only prefill/first-token and
never steady-state decode (\S\ref{sec:capsule})---so we do not characterize decode
rate here. The same contiguous-static-buffer, replay-whole design that yields this
TTFT floor is what makes the boundary state freezable
(\S\ref{sec:substrate})---the next layer.

\FloatBarrier
\subsection{Layer 2: the capsule mechanism}
\label{sec:eval-mech}

\paragraph{Capsule operation breakdown and scaling.} We isolate each capsule
operation and sweep the shared-prefix length on the long FP8-KV route (chunked
prefill, no per-position capture, so cold cost is genuine prefill compute), median
of $15$ repeats (Table~\ref{tab:break}).

\begin{table}[h]
\centering
\caption{Capsule cost breakdown vs.\ prefix length (paper TTFT: first base-logit
token, MTP draft-cache tail excluded; \S\ref{sec:eval-floor}). Snapshot, restore, and
the suffix append are all small and flat; cold grows with length, so the speedup
\emph{widens monotonically} $3.9\times\to27\times$.}
\label{tab:break}
\small
\begin{tabular}{@{}lcccccccc@{}}
\toprule
prefix & capsule & snapshot & restore & append & cold TTFT & capsule TTFT & speedup & tok-exact \\
\midrule
$2048$  & $160$\,MB & $0.3$\,ms & $0.3$\,ms & $25$\,ms & $\phantom{0}200$\,ms & $51$\,ms & $\phantom{0}3.92\times$ & yes \\
$4096$  & $224$\,MB & $0.4$\,ms & $0.4$\,ms & $26$\,ms & $\phantom{0}365$\,ms & $53$\,ms & $\phantom{0}6.91\times$ & yes \\
$8192$  & $352$\,MB & $0.7$\,ms & $0.7$\,ms & $28$\,ms & $\phantom{0}723$\,ms & $54$\,ms & $13.33\times$ & yes \\
$16384$ & $608$\,MB & $1.2$\,ms & $1.2$\,ms & $28$\,ms & $1541$\,ms & $57$\,ms & $\mathbf{26.94\times}$ & yes \\
\bottomrule
\end{tabular}
\end{table}

\noindent Snapshot and restore are \emph{sub-millisecond}---a bandwidth-bound copy
of the $160$--$608$\,MB capsule, scaling with size---and the suffix append is
\emph{flat at $\sim$$25$--$28$\,ms}, so the capsule TTFT stays flat (\,$51$--$57$\,ms)
while cold TTFT grows $200\to1541$\,ms over $2$k$\to$$16$k. The speedup therefore
\emph{widens monotonically with prefix length}, reaching $\mathbf{26.94\times}$ at
$16$k and continuing toward the $10$k--$50$k prefixes a coding agent resends each
turn. Output is token-exact versus a cold full prefill at every size. (Measuring TTFT
to the first base-logit token---excluding the decode-side MTP draft-cache tail
fill, \S\ref{sec:eval-floor}---is what makes the append flat; an earlier draft that
timed the whole prefill call counted that tail and saw a non-monotonic
append. A separate short, in-GPU route is also token-exact at $89.85$\,MB for a
$185$-token prefix; its cold includes per-position graph capture, so we do not read
its absolute TTFT as a prefill baseline.)

\paragraph{Not a better KV cache: the KV-only ablation.} The sharpest test that the
capsule is a \emph{whole-execution-state} object---not ``static buffers plus a KV
memcpy''---is to restore only what a positional KV cache structurally holds and show
it is not enough. We snapshot a $4096$-token boundary, then decode greedily ($K{=}0$)
three ways and compare to the full-restore reference (Table~\ref{tab:kvonly}). A
\emph{full} restore is token-exact. A \emph{KV-only} restore---keep the positional
full-attention KV, but drop the linear-attention recurrent$+$convolution fold (zero
it)---diverges at the \emph{first} token ($97.9\%$ of tokens mismatch). Restoring the
KV but leaving a \emph{stale} recurrent fold from an unrelated prompt diverges by the
third token ($93.8\%$). The recurrent state is a fold over the whole prefix with no
positional block (\S\ref{sec:bg}); positional KV reuse alone cannot reconstruct it, so a
reuse path built on positional KV alone is wrong here, not merely slow. This is the
capability the capsule adds, verified by divergence rather than asserted.

\begin{table}[h]
\centering
\caption{KV-only vs full-state capsule ($4096$-token boundary, greedy $K{=}0$, vs the
full-restore reference). Keeping the positional KV but dropping/staling the
recurrent$+$conv fold diverges almost immediately---the capsule's whole-state restore
is load-bearing, not a KV memcpy.}
\label{tab:kvonly}
\small
\begin{tabular}{@{}lccc@{}}
\toprule
restore variant & first divergence & tokens mismatched & token-exact \\
\midrule
full state (KV $+$ recurrent $+$ conv $+$ MTP) & --- & $0/48$ & \textbf{yes} \\
KV only, recurrent fold dropped (zeroed) & token $1$ & $47/48$ ($97.9\%$) & no \\
KV only, recurrent fold stale (other prompt) & token $3$ & $45/48$ ($93.8\%$) & no \\
\bottomrule
\end{tabular}
\end{table}

\FloatBarrier
\subsection{Layer 3: retention control vs an automatic cache}
\label{sec:eval-retention}

\begin{figure}[t]
\centering
\includegraphics[width=0.95\linewidth]{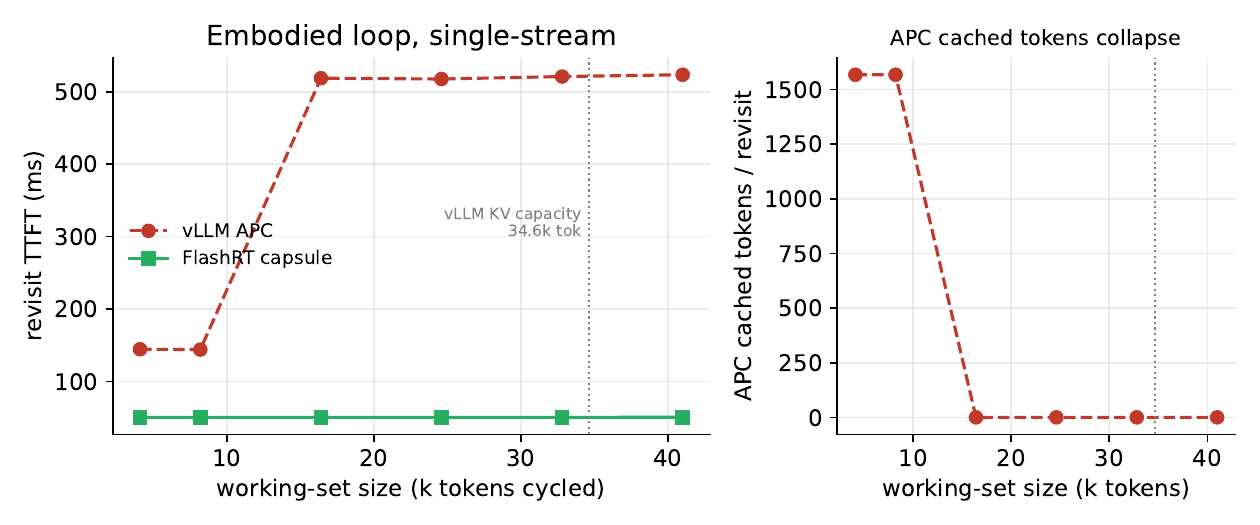}
\caption{Embodied loop, single-stream, under comparable \emph{reported device
residency} (capsule peak $27.5$\,GB torch-allocator vs vLLM $\sim$$30$\,GB reported
residency; different accounting semantics, no memory-win claim). \emph{Left:} an agent cycles $N$ distinct
$2048$-token contexts (skill/subgoal switch, interrupt-resume) and revisits each;
vLLM-APC reverts to cold prefill once reuse collapses, the capsule stays flat.
\emph{Right:} ground truth---APC's cached tokens per revisit drop $1568\to0$ at
$\approx 16$k tokens, \emph{below half} the reported $34.6$k full-attention KV
capacity, so raw KV capacity alone does not explain the collapse; in this hybrid run
the limiting path is the implementation's hybrid/recurrent prefix-cache retention.}
\label{fig:workingset}
\end{figure}

\paragraph{Versus vLLM on the same model and GPU.} We compare against
vLLM~$0.22.0$~\cite{paged2023} (torch~$2.11$/CUDA~$13$) at its best low-latency
config: NVFP4 (\code{compressed-tensors}), full CUDA-graph mode (not eager),
automatic prefix caching (APC) on, \code{max\_num\_seqs}$=1$,
\code{max\_model\_len}$=12288$, \code{gpu\_memory\_utilization}$=0.96$ (vLLM reports a
$35{,}746$-token GPU KV cache here; the working-set run in
Figure~\ref{fig:workingset} uses $0.95$, reporting $34{,}629$ tokens); the exact
command and engine logs are released with the code. TTFT is wall-clock to first token (\code{max\_tokens}$=1$); its
$\sim$$54$ tok/s single-stream decode confirms a healthy, un-crippled baseline. We
report \emph{absolute} TTFT, not speedup ratios (a slower system shows a larger
ratio). Table~\ref{tab:vllm}.

\begin{table}[h]
\centering
\caption{Absolute TTFT (ms) on the identical model/GPU (paper TTFT, MTP tail
excluded; vLLM has no MTP). No reuse: FlashRT cold is $2.6$--$2.8\times$ lower than
vLLM. With reuse: the capsule is $1.4$--$2.8\times$ lower than vLLM-APC.}
\label{tab:vllm}
\small
\begin{tabular}{@{}lcccc@{}}
\toprule
& \multicolumn{2}{c}{no reuse (cold)} & \multicolumn{2}{c}{with reuse} \\
\cmidrule(lr){2-3}\cmidrule(lr){4-5}
prefix & vLLM cold & FlashRT cold & vLLM APC & FlashRT capsule \\
\midrule
$2048$ & $\phantom{0}519$\,ms & $\mathbf{200}$\,ms & $143$\,ms & $\mathbf{51}$\,ms \\
$4096$ & $1026$\,ms & $\mathbf{365}$\,ms & $\phantom{0}76$\,ms & $\mathbf{53}$\,ms \\
$8192$ & $2057$\,ms & $\mathbf{723}$\,ms & $120$\,ms & $\mathbf{54}$\,ms \\
\bottomrule
\end{tabular}
\end{table}

\begin{figure}[t]
\centering
\includegraphics[width=0.62\linewidth]{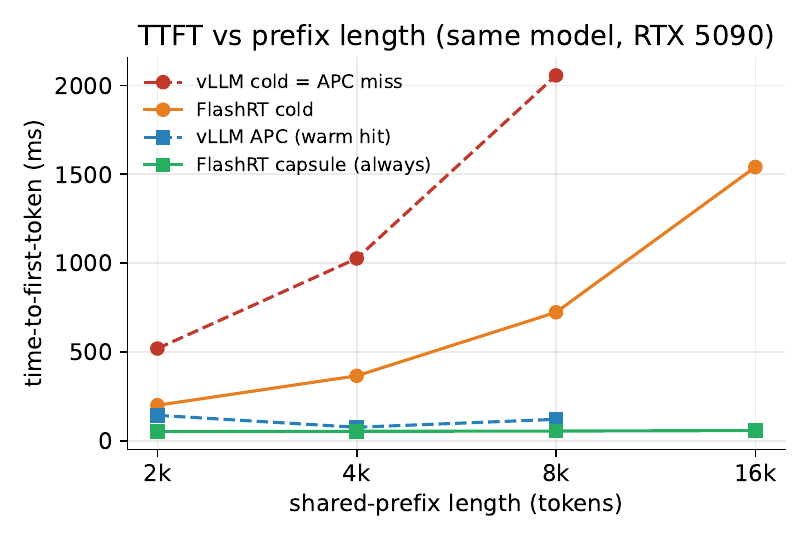}
\caption{Runtime floor and state-reuse floor, identical model/GPU, single-stream.
FlashRT \emph{cold} isolates the latency-first runtime substrate (the floor);
FlashRT \emph{capsule} adds the state mechanism that preserves it. vLLM \emph{APC}
is the best-case automatic KV-cache hit; vLLM \emph{cold} is its miss path (fresh
process, eviction, unseen prefix---common in the embodied regime). The capsule sits
\emph{below} the APC warm hit and holds its $\sim$$53$\,ms for any \emph{explicitly
pinned, GPU-resident} boundary (not dependent on automatic cache residency), extending
to $16$k where its speedup over cold is $27\times$. The vLLM comparison is reported up to $8$k under the
tested \code{max\_model\_len}$=12288$; the $16$k point is FlashRT cold-vs-capsule
scaling only.}
\label{fig:scaling}
\end{figure}

\noindent \textbf{The APC example---the crux of the mechanism argument.} vLLM's APC
is a top-tier prefix-reuse mechanism, and on the reuse it is built for it is fast.
Measured to the same first-base-logit-token TTFT (MTP excluded), the capsule's warm
reuse is nonetheless \emph{lower} than an APC hit ($51$/$53$/$54$ vs.\ $143$/$76$/$120$\,ms,
$1.4$--$2.8\times$): an APC hit still re-runs attention over the cached blocks and
re-prefills the non-cached remainder plus the suffix, whereas a capsule restores the
whole boundary as a ms-scale buffer copy and appends only the flat $\sim$$25$\,ms
suffix. But the magnitude is not the point we lean on; the structural point is that an
APC hit is \emph{opportunistic and cache-policy-controlled}---resident only while an
in-process, automatically LRU-managed, positional-only cache keeps the exact
prefix---whereas a capsule is an \emph{explicitly named, policy-pinned} execution
boundary. The miss modes we target are not a corner case for this regime---we model a
common physical-AI pattern (cycling skills/subgoals, interrupt-resume) below: a fresh
process, a restart, eviction under a small on-device budget (or the recurrent-retention
limit measured below), an intermittent control-loop session, a not-yet-seen branch.
There vLLM's TTFT is the \emph{cold} column ($519$--$2057$\,ms): \textbf{the cold
column \emph{is} APC's miss-path latency.} A \emph{GPU-resident} capsule holds its
$51$--$54$\,ms reuse latency for any pinned boundary ($\sim$$10$--$38\times$ below
vLLM's miss path; $27\times$ vs.\ cold at $16$k, Figure~\ref{fig:scaling});
host/disk-backed capsules add durability but pay a transfer on first promotion back to
the GPU tier (\S\ref{sec:capsule}). (Decode is not compared head-to-head: FlashRT uses
MTP speculative decode, this vLLM run does not---and the TTFT here excludes it on both
sides for fairness.)

\noindent The four numbers separate the two layers cleanly
(Table~\ref{tab:fourpaths}): the \emph{substrate} difference is the cold column
(latency-first runtime $200$ vs throughput-first runtime $519$\,ms), and the
\emph{mechanism} difference is within each runtime (reuse object vs none). The
capsule's reuse latency is held by explicit pinning, whereas the APC hit is held only
while the automatic cache still resides the prefix---so under the working set below,
vLLM falls back to its cold (miss) row while the capsule stays on its reuse row.

\begin{table}[h]
\centering
\caption{Four serving paths $=$ two substrates $\times$ \{no reuse, reuse object\},
at a $2048$-token prefix (ms, from Table~\ref{tab:vllm}). The capsule row is the
runtime floor \emph{preserved} by an explicitly pinned whole-boundary object.}
\label{tab:fourpaths}
\small
\begin{tabular}{@{}lllc@{}}
\toprule
path & runtime substrate & reuse object & TTFT \\
\midrule
vLLM cold        & throughput-first & none & $519$ \\
vLLM APC hit     & throughput-first & positional KV blocks (auto) & $143$ \\
FlashRT cold     & latency-first    & none & $200$ \\
\textbf{FlashRT capsule} & latency-first & \textbf{whole boundary (pinned)} & $\mathbf{51}$ \\
\bottomrule
\end{tabular}
\end{table}

\paragraph{The embodied loop, measured.} The miss path is not a corner case in this
regime: we model a common physical-AI pattern---an agent cycling among several
contexts (skills, subgoals, interrupt-resume), each with its own prefix. We cycle a
working set of $N$ distinct $2048$-token
contexts, revisit each single-stream, and read vLLM's per-request
\code{num\_cached\_tokens} as ground truth (Figure~\ref{fig:workingset}). vLLM-APC
reuses each context ($1568$ cached tokens---$2$ full $784$-token hybrid blocks of the
$2048$-token prefix; the partial third block is not a reusable full block---TTFT
$144$\,ms) up to a $\approx 8$k-token working set; beyond that its cached-token count
collapses to \emph{zero} and TTFT reverts to cold prefill ($519$\,ms). Critically, the
collapse occurs at $\approx 16$k tokens---\emph{less than half} of vLLM's own measured
$34{,}629$-token full-attention KV-cache capacity. The limiting factor in this
hybrid-mode APC run is therefore not raw full-attention KV capacity alone, but the
implementation's hybrid/recurrent prefix-cache retention path (an implementation-level
limit, marked experimental by vLLM---not a universal APC failure). The capsule store is explicit and persistable, so each
context's whole boundary stays pinned: TTFT is \emph{flat at $\approx 50$\,ms across
all $N$} (below even an APC hit), while holding all $20$ contexts ($3.4$\,GB of capsules) within a
$27.5$\,GB peak (torch allocator), comparable to vLLM's $\sim$$30$\,GB reported GPU
residency---so the flat latency holds under \emph{comparable reported device
residency}, despite different accounting semantics. The two frameworks measure memory
differently (torch allocator vs vLLM's reported KV/reserved pool); we release both
numbers and make \emph{no} memory-budget win claim.
vLLM offers opt-in CPU/disk KV-offload tiers that would raise the threshold, but they
remain automatic, block-level, positional-only, and add transfer latency; they are
not explicit per-context pinning, a session snapshot, fork, or rollback
(Table~\ref{tab:cap}).

\paragraph{What block/radix caches do not make first-class.} The differentiation
is not TTFT magnitude but \emph{what} is a reusable object. A KV cache can share
positional KV, but it does not expose a whole-boundary snapshot that includes the
recurrent, convolution, MTP, and graph-bound state---supporting that would require an
additional state-snapshot mechanism outside the KV/radix object. Concretely, in both
baselines the recurrent state is handled by a \emph{separate, special-cased mamba
cache}, not the prefix cache: vLLM logs that ``prefix caching in Mamba cache `align'
mode is\ldots experimental'' and forces the $784$-token block above; SGLang's
\code{qwen3\_next} attaches RadixAttention only to the full-attention layers and keeps
the gated-delta-net state in a separate cache. The capsule makes the entire boundary
one explicit object and restores it as a single byte-exact snapshot, verified by
token-exact end-to-end behavior (Table~\ref{tab:kvonly},~\ref{tab:cap}).

\begin{table}[h]
\centering
\caption{Mechanism comparison: \emph{what} can be reused, and \emph{how} that reuse
is controlled and persisted. ``No'' means the KV/radix managed object does not expose
this as a first-class operation, not that the codebase could never be extended with an
additional snapshot system (\S\ref{sec:bg}). $^\ast$vLLM's hybrid prefix caching is marked
experimental and forces a $784$-token attention block. $^\dagger$Capsule bytes
persist to host/disk, but the latency floor returns only after the runtime/graph is
rebuilt and the capsule is promoted to the GPU tier (not a cross-restart warm
floor). $^\ddagger$Recurrent/conv state is restored within the byte snapshot and
verified by end-to-end token-exactness (Table~\ref{tab:kvonly}), not a per-buffer
numerical compare.}
\label{tab:cap}
\footnotesize
\setlength{\tabcolsep}{4pt}
\begin{tabular}{@{}p{0.40\linewidth}*{3}{>{\centering\arraybackslash}p{0.155\linewidth}}@{}}
\toprule
 & vLLM APC & SGLang radix & FlashRT capsule \\
\midrule
\emph{what is reused} & & & \\
\quad positional (full-attn) KV & yes & yes & yes \\
\quad linear-attn recurrent state & exp.$^\ast$ & sep.\ cache & \textbf{yes}$^\ddagger$ \\
\quad conv state & sep.\ cache & sep.\ cache & \textbf{yes}$^\ddagger$ \\
\emph{how it is controlled} & & & \\
\quad whole-boundary fork ($1\!\to\!N$, incl.\ recurrent/conv) & no & no & \textbf{yes} \\
\quad rollback to an earlier whole boundary & no & no & \textbf{yes} \\
\quad explicit pin / what-to-keep & auto LRU & auto LRU & \textbf{policy} \\
\quad session \emph{bytes} persist across restart$^\dagger$ & no & no & \textbf{yes} \\
\bottomrule
\end{tabular}
\end{table}

\FloatBarrier
\subsection{Cross-domain: one contract}
\label{sec:eval-cross}

\begin{figure}[t]
\centering
\includegraphics[width=0.5\linewidth]{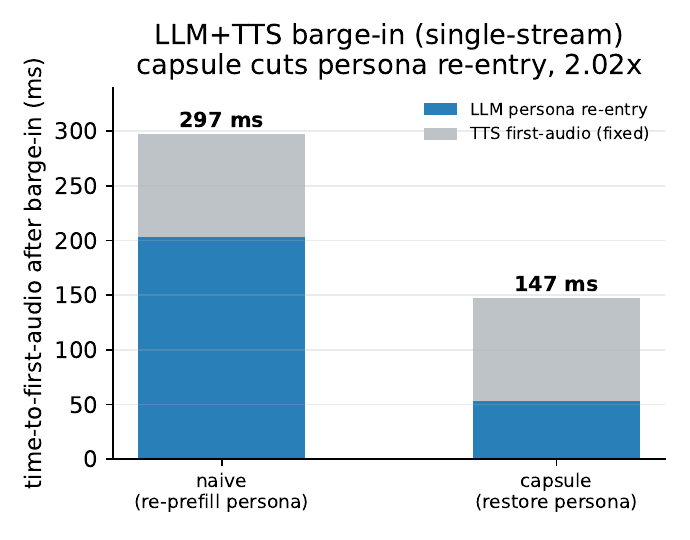}
\caption{LLM+TTS barge-in, single-stream, \emph{composed} latency: measured LLM
persona re-entry $+$ a separately measured fixed TTS first-audio term ($94$\,ms),
\emph{not} a co-resident end-to-end measurement. After an interrupt, \emph{capsule}
(restore the pinned persona) cuts the LLM re-entry vs \emph{naive} (re-prefill);
the TTS term is identical for both, so the capsule is the $2.02\times$ difference.}
\label{fig:bargein}
\end{figure}

\paragraph{Multi-model interactive serving: LLM+TTS barge-in (composed estimate).}
The capsule mechanism spans model types. A voice assistant carries a fixed
persona/system prefix; the user \emph{barges in} with a new instruction. We report a
\emph{composed} barge-in latency---\textbf{not} a co-resident end-to-end pipeline
measurement: we measure the LLM persona re-entry (Qwen3.6) and add a separately
measured fixed TTS first-audio term ($94$\,ms, the fp8 number below; co-loading 27B
LLM $+$ TTS exceeds the GPU at the long-route \code{max\_seq}). The two ways to
handle the LLM persona re-entry (Figure~\ref{fig:bargein}): \emph{naive} re-prefills
the $2048$-token persona from cold; \emph{capsule} restores the pinned persona and
appends only the new instruction. Since the TTS term is identical for both arms, the
difference is exactly the LLM persona re-entry: the capsule cuts it from $\sim$$203$
to $\sim$$53$\,ms, so the composed TTFA-after-barge-in is $147$ vs.\ $297$\,ms
($\mathbf{2.02\times}$); the capsule arm is flat ($145$--$148$\,ms) while the naive
arm varies with re-prefill ($296$--$369$\,ms). This is the same restore-not-recompute
advantage as the LLM working-set, in an interactive two-model framing. (One-context
co-hosting itself is the contract mechanism validated by the robot hand-off,
\S\ref{sec:eval}; a co-resident end-to-end pipeline with streaming overlap is future
work.)

\begin{figure}[t]
\centering
\includegraphics[width=0.92\linewidth]{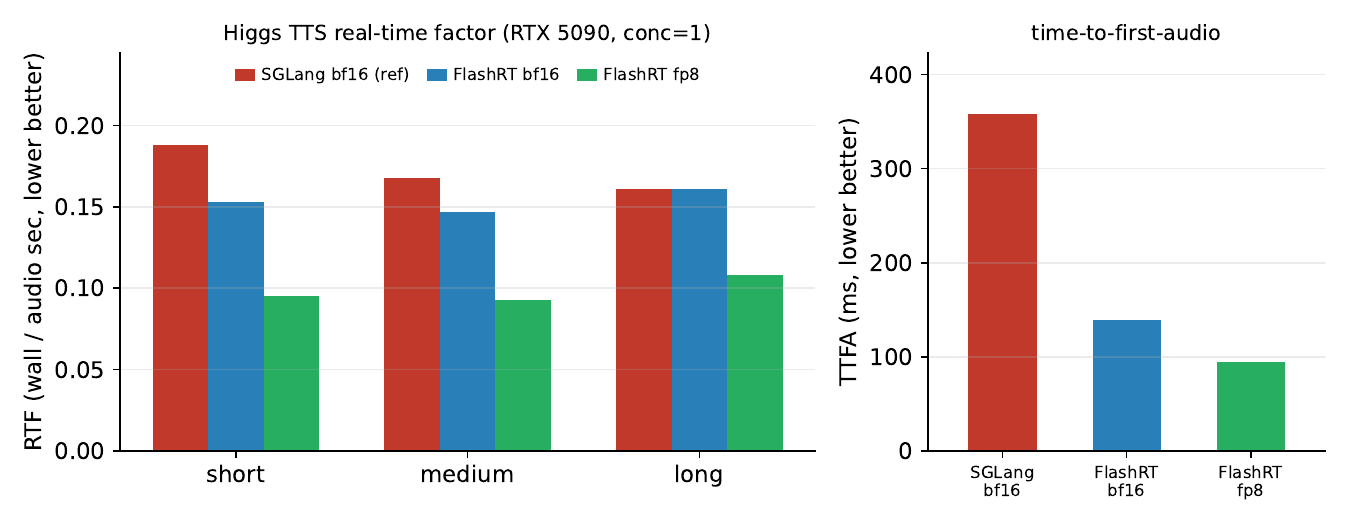}
\caption{Higgs TTS, RTX 5090, single-stream (concurrency $1$). Left: real-time
factor (lower better). Right: time-to-first-audio. At matched precision (bf16)
FlashRT's TTFA is $2.6\times$ lower than sglang-omni and per-frame RTF is on par;
fp8 adds a further speedup. The edge is low-overhead execution, not a faster kernel.}
\label{fig:higgs}
\end{figure}

\paragraph{Runtime sanity check on a SGLang-native model: Higgs TTS.}
To test the same single-stream regime against SGLang where it is the well-tuned,
\emph{native} path, we run Higgs Audio v3 TTS (4B; its architecture ships only in
sglang-omni, which is therefore the canonical reference), RTX 5090, concurrency
$1$ (Figure~\ref{fig:higgs}). The user-facing metric is \emph{time-to-first-audio}
(TTFA): at \emph{matched precision} (bf16), FlashRT's TTFA is $139$\,ms versus
sglang-omni's $\sim$$358$\,ms ($2.6\times$ lower), and our fp8 path reaches
$94$\,ms ($3.8\times$); streaming output is cosine $1.0$ vs.\ one-shot. Steady-state
real-time factor is on par at matched precision (RTF $0.161$ vs.\ $0.161$ long) and
$1.6$--$1.8\times$ better in fp8. Crucially, the matched-precision per-frame compute
is comparable---FlashRT's responsiveness edge is \emph{clean, low-overhead
execution and tight chunking, not a faster kernel}. (FlashRT also resides in
$6.6$--$10.3$\,GB vs.\ $28.3$\,GB, since sglang reserves a KV/batch pool unused at
concurrency $1$---relevant for on-device deployment, but not the point here.) This
experiment does \emph{not} evaluate capsules; it checks that the same latency-first
execution contract also helps a different single-stream workload---supporting
evidence for the runtime, not for the capsule mechanism.

\paragraph{Robot side (byte-identical action replay).} On an FP8 advantage-conditioned
(classifier-free-guidance) $\pi_0$-style diffusion policy~\cite{pi0_2024}, the
rollout boundary is snapshotted into a $640$-byte contract Buffer and restored;
replaying the captured policy graph reproduces the action \emph{byte-for-byte}
(cosine $1.000000$), including after the live boundary buffer is deliberately
overwritten by an unrelated episode. The capsule here is
literally one \code{frt} Buffer and a device-to-device copy---no model-specific
API---so episode reset is the same verb as the LLM capsule. The same contract also
drives VLA \emph{inference} usability for exactly the state changes this regime
demands: in a planner$\to$actor hand-off (two co-hosted Pi05 policies, one exec
context, $1{:}4$ multi-rate), a mid-run \emph{interrupt} that injects a new subgoal
overwrites a shared buffer and the next actor tick consumes it with \emph{no graph
recapture} (verified). These are \emph{offline policy-graph mechanism tests}, not
on-robot success-rate results: they establish that episode reset, interrupt, and
subgoal injection are correct, zero-recapture contract operations. Quantified
on-robot rollout latency, task success, and a safety study are future work.

\FloatBarrier
\subsection{On-device replication: Jetson AGX Thor (SM110) and DGX Spark (GB10)}
\label{sec:eval-thor}

The regime's defining target is the edge, so we replicate the LLM and robot-policy
mechanism evidence on two real on-device systems, both aarch64 with unified memory.
\paragraph{Jetson AGX Thor (\code{sm\_110}).} An NVIDIA Jetson AGX Thor (aarch64, \code{sm\_110}, CUDA
$13.1$, $122$\,GiB unified LPDDR5X), same Qwen3.6 NVFP4-W4A16 model, single-stream,
same paper-TTFT convention (MTP tail excluded). Every claim that held on the 5090
holds here, and the headline transfers \emph{stronger}. The capsule is \emph{correct}
(token/byte-exact incl.\ the chunk-alignment condition, plus the adversarial
dirtied-state test), \emph{tight-tailed} (at $4$k, cold p99 within $\sim$$1$\,ms of p50,
capsule within $\sim$$0.5$\,ms), and a \emph{whole-execution-state} object (KV-only
restore diverges $97.9\%$ at the first token---the recurrent fold is load-bearing on
SM110 too). Cold prefill on the edge device costs \emph{seconds} ($2.2$--$17.9$\,s over
$2$k--$16$k), while capsule restore is a $2.5$--$13$\,ms buffer copy and the append is
flat ($\sim$$150$\,ms), so the cold$\to$capsule speedup is $9$--$76\times$, \emph{wider}
than the 5090's $27\times$ because the eliminated cold cost is larger.

Against vLLM~$0.23.0$ on the same Thor (NVFP4-W4A16 via vLLM's Marlin FP4-weight
kernel---Thor has no native FP4, vLLM's own warning; APC on, \code{max\_num\_seqs}$=1$),
absolute within-device TTFT, FlashRT wins \emph{all eight cells} (Table~\ref{tab:thor}):
cold is $1.15$--$1.21\times$ below vLLM's cold, and the capsule is $1.5$--$4.4\times$
below a vLLM-APC hit, \emph{widening} with prefix. Robot (T4) episode reset is
byte-identical (cos $1.0$) through the exec contract, planner$\to$actor hand-off with a
mid-run interrupt runs with no recapture (T5), and \emph{fork} and \emph{rollback} are
token-exact (T9, below). One honest difference from the 5090: vLLM-APC does \emph{not}
collapse on Thor (capsule flat $173$\,ms vs APC flat $664$\,ms to a $41$k working set),
because Thor's large unified memory gives a $\sim$$474$k-token KV with no eviction
pressure---so the APC-collapse contrast (Fig.~\ref{fig:workingset}) is a 5090
discrete-VRAM effect, and the Thor reuse win is simply the lower absolute capsule TTFT.
(FlashRT's MTP decode kernel on Thor is unoptimized; decode stays out of scope, the
axis is TTFT.)

\begin{table}[h]
\centering
\caption{Jetson AGX Thor (SM110), same model, within-device absolute TTFT (ms),
paper-TTFT convention. FlashRT is lower in all eight cells; cold$\to$capsule speedup
$9$--$76\times$ (wider than the 5090's because Thor's cold prefill is in seconds).}
\label{tab:thor}
\small
\begin{tabular}{@{}lcccc@{}}
\toprule
& \multicolumn{2}{c}{no reuse (cold)} & \multicolumn{2}{c}{with reuse} \\
\cmidrule(lr){2-3}\cmidrule(lr){4-5}
prefix & vLLM cold & FlashRT cold & vLLM APC & FlashRT capsule \\
\midrule
$2048$  & $\phantom{0}2529$ & $\mathbf{2196}$ & $\phantom{0}663$ & $\mathbf{240}$ \\
$4096$  & $\phantom{0}5197$ & $\mathbf{4208}$ & $\phantom{0}332$ & $\mathbf{225}$ \\
$8192$  & $10121$ & $\mathbf{8502}$ & $\phantom{0}586$ & $\mathbf{244}$ \\
$16384$ & $20748$ & $\mathbf{17922}$ & $1036$ & $\mathbf{236}$ \\
\bottomrule
\end{tabular}
\end{table}

\begin{figure}[h]
\centering
\includegraphics[width=0.95\linewidth]{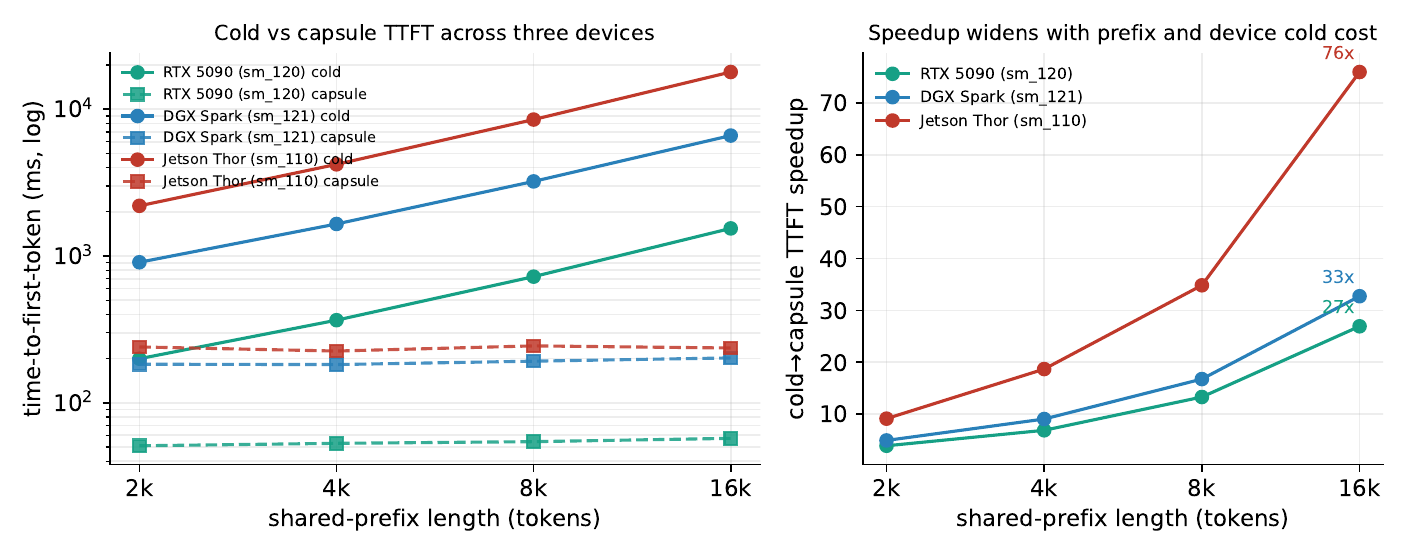}
\caption{Cross-device, single-stream, paper-TTFT convention (first base-logit token,
MTP tail excluded). \emph{Left:} cold vs capsule TTFT vs prefix length on all three
devices (log scale)---cold prefill rises with prefix and is ordered by device compute
(Thor $>$ Spark $>$ 5090), while the capsule is flat and clustered low ($\sim$$50$--$240$\,ms)
on every device. \emph{Right:} the cold$\to$capsule speedup widens with prefix and tracks
each device's recompute cost ($27\times$ on the 5090, $33\times$ on Spark, $76\times$ on Thor
at $16$k). The capsule TTFT itself stays $\sim$$0.05$--$0.24$\,s independent of prefix and
architecture; only the eliminated cold cost differs.}
\label{fig:crossdev}
\end{figure}

\begin{figure}[h]
\centering
\includegraphics[width=0.72\linewidth]{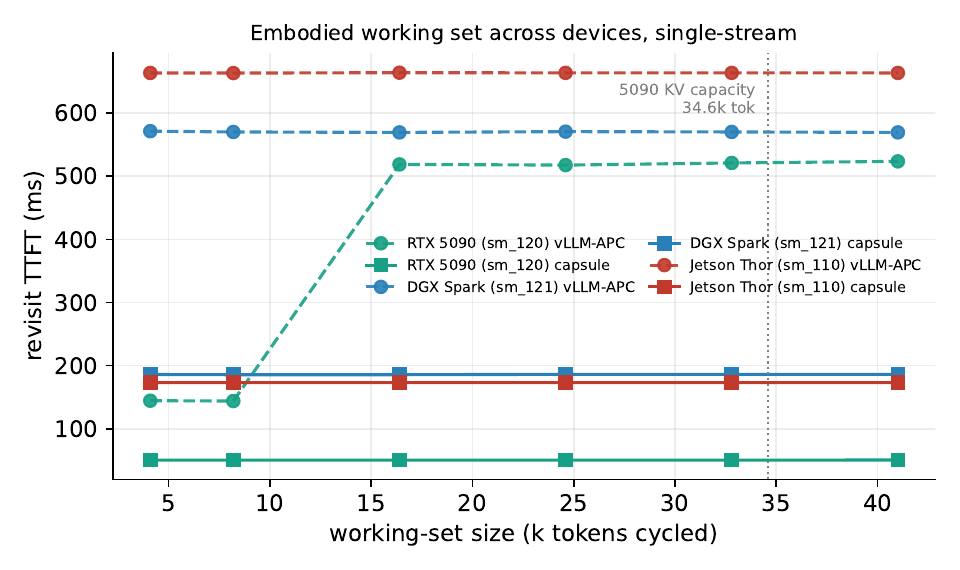}
\caption{Embodied working set across the three devices, single-stream: revisit TTFT as
$N$ distinct $2048$-token contexts are cycled (working set $4.1$--$41$k tokens). The
\emph{capsule} (solid) is flat on every device ($\sim$$50$/$173$/$186$\,ms on
5090/Thor/Spark). \emph{vLLM-APC} (dashed) collapses to a cold miss only on the 5090,
whose discrete VRAM evicts past its $34.6$k-token KV capacity; on Thor and Spark the large
unified memory keeps APC flat (but still $3$--$4\times$ above the capsule). Under this
pinned working set, capsule revisit latency is set by explicit residency rather than
automatic cache eviction, whereas APC retention is device-dependent.}
\label{fig:crossdev-ws}
\end{figure}

\paragraph{DGX Spark (GB10, \code{sm\_121}).} A second unified-memory on-device system---an
NVIDIA DGX Spark (GB10 Grace--Blackwell, aarch64, capability $(12,1)$, $121$\,GiB unified
LPDDR5X; same Qwen3.6 NVFP4-W4A16 model, single-stream, same paper-TTFT convention)---reproduces
every property. Because its capability is not $(11,0)$, the engine auto-selects the same
Blackwell frontend as the 5090 and the capsule API is inherited unchanged. The gate is
token/byte-exact (incl.\ the chunk-alignment condition and the adversarial dirtied-state
test); the tail is tight (at $4$k, cold p99 within $0.5\%$ of p50, capsule within $1.2\%$);
and a KV-only restore diverges at the first token while the full capsule is token-exact
(whole-execution-state). Cold TTFT grows with prefix ($0.9$--$6.6$\,s over $2$k--$16$k) while
the capsule is flat ($\sim$$183$--$202$\,ms; restore $1.8$--$6.6$\,ms, append $\sim$$0.1$\,s),
so the cold$\to$capsule speedup is $5$--$33\times$---smaller than Thor's only because Spark's
faster GPU makes the eliminated cold cost smaller; the capsule TTFT itself is the same
$\sim$$0.2$\,s. Against vLLM~$0.23.0$ (again the Marlin FP4-weight fallback---no native FP4 on
\code{sm\_121} either; APC on, \code{max\_num\_seqs}$=1$), FlashRT cold is $2.3$--$2.5\times$
below vLLM cold and the capsule is $1.7$--$4.2\times$ below a vLLM-APC hit at every prefix
(Table~\ref{tab:spark}); \emph{fork} and \emph{rollback} are token-exact. As on Thor,
vLLM-APC does not collapse out to a $41$k-token working set (the large unified KV has no
eviction pressure), so the reuse win is again the lower absolute capsule TTFT. The mechanism
thus holds identically across three architectures---\code{sm\_120} (5090), \code{sm\_110}
(Thor), \code{sm\_121} (Spark)---and the cold$\to$capsule multiplier simply tracks each
device's recompute cost while the capsule TTFT stays $\sim$$0.2$\,s (Fig.~\ref{fig:crossdev}).
The embodied working-set picture is likewise consistent across devices
(Fig.~\ref{fig:crossdev-ws}): the capsule revisit TTFT is flat as $N$ distinct $2$k-token
contexts are cycled out to a $41$k working set on all three (5090 $\sim$$50$\,ms, Thor
$\sim$$173$\,ms, Spark $\sim$$186$\,ms; $20$ pinned Spark capsules use $3.35$\,GB), whereas
vLLM-APC collapses to a cold miss only on the 5090 (its discrete VRAM evicts past the
$34.6$k-token KV) and stays flat-but-higher on Thor/Spark, whose large unified memory has
no eviction pressure---so APC retention is device-dependent while capsule retention is not.

\begin{table}[h]
\centering
\caption{NVIDIA DGX Spark (GB10, \code{sm\_121}), same model, within-device absolute TTFT
(ms), paper-TTFT convention. FlashRT is lower in all eight cells; cold$\to$capsule speedup
$5$--$33\times$. vLLM-APC is non-monotonic in prefix (at $2$k only part of the prefix is
cached, so more is re-prefilled); the capsule TTFT is flat regardless.}
\label{tab:spark}
\small
\begin{tabular}{@{}lcccc@{}}
\toprule
& \multicolumn{2}{c}{no reuse (cold)} & \multicolumn{2}{c}{with reuse} \\
\cmidrule(lr){2-3}\cmidrule(lr){4-5}
prefix & vLLM cold & FlashRT cold & vLLM APC & FlashRT capsule \\
\midrule
$2048$  & $\phantom{0}2122$ & $\mathbf{\phantom{0}906}$ & $571$ & $\mathbf{183}$ \\
$4096$  & $\phantom{0}4056$ & $\mathbf{1652}$ & $302$ & $\mathbf{182}$ \\
$8192$  & $\phantom{0}7935$ & $\mathbf{3221}$ & $482$ & $\mathbf{192}$ \\
$16384$ & $16129$ & $\mathbf{6613}$ & $839$ & $\mathbf{202}$ \\
\bottomrule
\end{tabular}
\end{table}

\FloatBarrier
\section{Scope and Limitations}
\label{sec:scope}

We are deliberately explicit about boundaries.
\begin{itemize}[leftmargin=1.4em,itemsep=2pt]
\item \textbf{A capsule is a binary state blob} bound to exact weights,
quantization, kernel version, and graph bucketing. Persistence (disk, shared
store) is for warm-starting the \emph{same} deployment or sharing within a team---
not a portable, cross-version, token-level text cache.
\item \textbf{Static-buffer graph-plan capture trades flexibility for latency.} It
requires a bounded set of shape variants (the ShapeKey table with LRU eviction) and a
fixed maximum sequence length; this suits low-concurrency, few-shape workloads and is
\emph{ill-suited} to many concurrent, highly variable shapes, where paged/radix
engines are the right tool. The capsule lives on the latency-first side of that
trade by design, not by oversight.
\item \textbf{Single node, latency-first, low concurrency.} The registry tiers
naturally (GPU $\to$ host RAM $\to$ disk) but is single-node; large-cluster
distributed KV is intentionally out of scope. We make no high-concurrency
throughput claim and do not compete with paged/radix engines on their main turf.
\item \textbf{Capsules are opt-in} and additive: they do not change the execution
contract beyond the one mechanism of \S\ref{sec:contract}, and do not change
steady-state decode.
\item \textbf{Production agent integration is ongoing.} A controlled microbenchmark
isolates the capsule mechanism; a full multi-turn server (automatic prefix
matching, OpenAI-style full-history clients, graph-cache robustness) has open
engineering issues and is future work.
\item \textbf{On-robot evaluation is future work}, by us and collaborating
institutions. This paper establishes the mechanism and its correctness (byte-exact
stored state, token/action-identical output),
not a robotics result.
\item \textbf{Baselines.} We benchmark vLLM directly on the same hybrid model. We
use SGLang primarily as the representative of the \emph{radix-prefix managed object},
not as a same-model latency baseline: it serves this arch but, like vLLM,
special-cases the recurrent state outside its radix cache, so the capability gap is
established from source (\S\ref{sec:eval}). The Higgs TTS numbers are a single-stream
\emph{runtime} sanity check on a model SGLang serves natively, not a substitute for a
same-model capsule comparison; a same-model SGLang latency race is deferred to a
follow-up.
\item \textbf{The long ``TQ'' KV mode} is not yet wired; \code{snapshot\_capsule()}
fails loudly rather than producing a partial capsule.
\end{itemize}

\FloatBarrier
\section{Related Work}
\label{sec:related}

\textbf{Paged and radix KV caches.} PagedAttention/vLLM~\cite{paged2023} manage KV
memory as OS-style pages with copy-on-write sharing and automatic prefix caching;
RadixAttention/SGLang~\cite{sglang2024} reuse longest common prefixes via a radix
tree with cache-aware scheduling. Both manage a positional KV fragment under
eager/piecewise execution; capsules manage the whole graph-bound execution state under
static-buffer graph-plan capture, and additionally express fork, rollback, and hybrid
recurrent-state reuse. \textbf{Contiguous-KV alternatives.} vAttention~\cite{vattention2025} keeps
KV virtually contiguous via CUDA virtual memory rather than paging, sharing our
preference for contiguity but still managing only the KV cache, not the whole
graph-bound execution state. \textbf{Stateful serving and shared prefixes.} Unlike process
checkpoint/restore~\cite{criu}, a capsule is not a full process snapshot but the
model-specific graph-bound continuation state needed for the next replay; and unlike
shared-prefix mechanisms---Pensieve~\cite{pensieve2025} caches multi-turn conversation
state across requests, Hydragen~\cite{hydragen2024} and Prompt
Cache~\cite{promptcache2024} reuse shared-prefix attention/KV state---the reused object
is not KV-derived prefix state but the closed graph-bound buffer set (recurrent/conv
included), and capsules add fork/rollback.
\textbf{CUDA Graphs}~\cite{cudagraphs} are the low-latency primitive our static-buffer
graph capture builds on; mainstream serving uses them while keeping KV out of the graph,
which is the distinction we draw in \S\ref{sec:bg}. \textbf{Hybrid /
linear-attention models}~\cite{mamba2023,gateddeltanet2024} have a recurrent state
that is a fold over the whole prefix; we show its prefix reuse must go through a
state snapshot, with a chunk-alignment condition for exactness.
\textbf{Checkpoint/restore and VM snapshots}~\cite{criu} inspire the verbs;
capsules bring freeze/restore/fork to the execution state of a captured-graph
inference. \textbf{VLA policies.} $\pi_0$-style flow/diffusion
policies~\cite{pi0_2024,flowmatching2022} are the robot models we drive; capsules
provide their episode reset and warm re-entry.

\FloatBarrier
\section{Conclusion}
\label{sec:conclusion}

Treating an inference session as a checkpointable, forkable object---rather than a
stream over a paged cache---gives a third managed object for serving systems. By
capturing the whole forward as a graph plan over contiguous static buffers, FlashRT
makes the committed state a fixed buffer set; freezing it into an execution-state
capsule turns prefix reuse into a bandwidth-bound copy-and-rebuild, makes fork and
rollback first-class operations through the same copy-and-restore primitive, and
unifies---at the execution-mechanism level---an LLM agent's warm start with a robot
rollout's episode reset under one mechanism. Restore is byte-exact for the stored state and token-exact for
the tested LLM paths, and the TTFT speedup widens with prefix length. The graph
decides how to compute; the capsule decides which state to compute from; each step
pays only for the current input.

\appendix

\FloatBarrier
\section{Same-model SGLang latency: attempt and fairness caveat}
\label{app:sglang}

For completeness we attempted a same-hybrid-model SGLang latency point to sit beside
the vLLM numbers (\S\ref{sec:eval-retention}), in a dedicated environment (SGLang
$0.5.13$, transformers $5.8.1$, \code{sgl\_kernel}$+$\code{flashinfer}, same RTX~5090).
The architecture \emph{is} supported: transformers recognizes
\code{Qwen3\_5ForConditionalGeneration} and SGLang ships a matching model
(\code{models/qwen3\_5.py}), so the engine loads the config and \emph{begins building
the model}. It then aborts inside the \code{compressed-tensors} quantization path:
our checkpoint is NVFP4 \emph{weight-only} (W4A16---\code{weights}: $4$-bit float,
\code{input\_activations}: \code{None}), and SGLang $0.5.13$'s scheme detector has no
weight-only-NVFP4 case and dereferences the absent activation-quant
(\code{AttributeError: 'NoneType' \ldots num\_bits} in \code{\_is\_static\_tensor\_w8a8}).
vLLM serves exactly this scheme (\S\ref{sec:eval-retention}); this SGLang release does
not. A clean number therefore awaits either a SGLang release that supports NVFP4-W4A16
\code{compressed-tensors} or a re-export of the model to a SGLang-supported scheme
(which would no longer be the identical checkpoint). We deliberately do \emph{not} work
around it by patching SGLang, rewriting the checkpoint's quantization, or changing its
\code{model\_type}, since any of these yields a non-canonical, unfair configuration. We
therefore report SGLang \emph{structurally} (radix-prefix managed object, from source,
Table~\ref{tab:cap}) and \emph{experimentally} only through the SGLang-native Higgs-TTS
runtime check (\S\ref{sec:eval-cross}); a same-model SGLang latency number is left to
the multi-hardware follow-up. The attempt is fully reproducible:
\code{repro/sglang\_qwen36\_ttft.py} with the environment and traceback recorded in the
released artifacts.

{\small
\bibliographystyle{plain}
\bibliography{refs}
}

\end{document}